\pdfoutput=1

\documentclass[11pt]{article}

\usepackage[]{emnlp2021}

\usepackage{times}
\usepackage{makecell}
\usepackage{latexsym}

\usepackage[T1]{fontenc}

\usepackage[utf8]{inputenc}
\usepackage{multirow}
\usepackage{booktabs}
\usepackage{graphicx}
\usepackage{amsmath}
\usepackage{svg}
\usepackage{enumitem}
\usepackage{microtype}

\definecolor{myyellow}{RGB}{253, 231, 31}
\definecolor{mypurple}{RGB}{68, 1, 84}

\title{Does Knowledge Help General NLU? An Empirical Study}

\author{Ruochen Xu\thanks{$\;\;$Equal contribution}$\;$, Yuwei Fang$^*$, Chenguang Zhu, Michael Zeng \\
Microsoft Cognitive Services Research Group\\
  \texttt{\{ruox,yuwfan,chezhu,nzeng\}@microsoft.com} \\}

\begin{document}
\maketitle
\begin{abstract}
It is often observed in knowledge-centric tasks (e.g., common sense question and answering, relation classification) that the integration of external knowledge such as entity representation into language models can help provide useful information to boost the performance. 
However, it is still unclear whether this benefit can extend to general natural language understanding (NLU) tasks. In this work, we empirically investigated the contribution of external knowledge by measuring the end-to-end performance of language models with various knowledge integration methods. We find that the introduction of knowledge can significantly improve the results on certain tasks while having no adverse effects on other tasks. We then employ mutual information to reflect the difference brought by knowledge and a neural interpretation model to reveal how a language model utilizes external knowledge. Our study provides valuable insights and guidance for practitioners to equip NLP models with knowledge. 
\end{abstract}

\section{Introduction}
Language models utilize contextualized word representations to boost the performance of various NLP tasks \cite{devlin2019bert,liu2019roberta,Lan2020ALBERT:,Clark2020ELECTRA:}. In recent years, there has been a rise in the trend of integrating external knowledge into language models \cite{wang2020k,yu2020jaket,wang2019kepler,liu2020k,peters2019knowledge,zhang2019ernie,e_bert} based on the transformer \cite{transformer}. For instance, representations of entities in the input and related concepts are combined with contextual representations to provide additional information, leading to significant improvement in many tasks \cite{han2018fewrel,choi2018openentity,ling2012FIGER,berant2013webquestion}. 

However, most of these approaches \cite{zhang2017TACRED,talmor2019commonsenseqa} focus on knowledge-centric tasks, e.g. common sense Q\&A, relation extraction, where the completion of a task requires information from an external source other than the input text. But, these works overlook many more general NLP tasks which do not explicitly request usage of knowledge, including but not limited to sentiment classification, natural language inference, sentence similarity, part-of-speech tagging, and named entity recognition \cite{glue,superglue}.
So far, the improvement on these tasks often originates from more sophisticated architectures, larger model size, and an increasing amount of pre-training data.
Very little work has investigated whether external knowledge will improve the performance of these non-knowledge-centric tasks.

In this work, we aim to find out whether external knowledge can lead to better language understanding ability for general NLU tasks. Specifically, we want to answer the following questions:
\begin{itemize}[leftmargin=*]
  \setlength\itemsep{0.18em}
    \item First of all, does knowledge help general NLU tasks overall? (Section~\ref{sec:kg_results} Q1)
    \item Among various NLU tasks, what source of knowledge and which tasks could benefit the most from the integration of external knowledge? (Section~\ref{sec:kg_results} Q2)
    \item Under the same experimental settings, which integration methods are the most effective in combining knowledge with language models? (Section~\ref{sec:kg_results} Q3)
    \item Which large-scale pre-trained language models benefit the most from external knowledge? (Section~\ref{sec:kg_results} Q4)
    \item Can we tell knowledge is helpful besides the end-to-end performance indicator? (Section~\ref{sec:analysis} Q1)
    \item If knowledge can help with certain NLU tasks, how does the language model utilize the external knowledge? (Section~\ref{sec:analysis} Q2)
\end{itemize}

Answers to these questions not only help us understand how knowledge is leveraged in language models but also provide important insights into how to leverage knowledge in various NLU tasks.

In detail, we explore the different sources of knowledge including the textual explanation of entities and their embeddings. We explore two main categories of knowledge integration methods. \textbf{Knowledge as Text} places descriptions of entities into the input text, with normal or modified attention mechanism from the language model. \textbf{Knowledge as Embedding} integrates contextual or graphical embedding of entities into the language model via addition. Both methods are non-invasive, meaning that the language model's inner structure does not need to be altered.
We apply these knowledge integration methods to 4 pretrained language models and conduct extensive experiments on 10 NLU tasks. The results show that introducing knowledge can outperform vanilla pretrained language models by 0.46 points by averaging all language models.

To understand how and why knowledge integration methods can help with language models, we also utilize mutual information (MI) to reflect the difference brought by knowledge (see Figure~\ref{fig:ana_mi}) and visualize the contribution of inputs to the prediction of knowledge-enhanced language models (see figure \ref{fig:diffmask-cola-exmp}) to better understand the interaction between language model and knowledge.
We find that $(i)$ Knowledge integration methods retain more information about input while gradually discard take-irrelevant information and finally keep more information about output.
$(ii)$ Although the knowledge is only introduced for a subset of tokens in the input sentence, it affects the decision process of the model on all tokens and improves the generalization ability on certain tasks.


In summary, we present a systematic empirical analysis on how to effectively integrate external knowledge into existing language models for general NLU tasks. This provides valuable insights and guidance for practitioners to effectively equip language models with knowledge for different NLU tasks.


\section{Related Work}
\label{sec:related-work}
\label{rel:kg_combine}
In this section, we review previous works that explore how to combine external knowledge with language models, which can be grouped into the following categories.  

\textbf{Joint Pretraining} 
Some recent works combine pre-trained language models with external knowledge by joint pretraining with both unstructured text and structured knowledge bases.
Ernie(Baidu) \cite{sun2019ernie} modified pretrain objective in BERT \cite{devlin2019bert} to mask the whole span of named entities. WKLM~\cite{Xiong2020Pretrained} trains the model to detect whether an entity is replaced by another one in the same category.
LUKE~\cite{yamada2020luke} propose a pretrained model which uses similar entity masks from Wikipedia in pretraining but treats words and entities in a given text as independent tokens.
KEPLER~\cite{wang2019kepler} and JAKET~\cite{yu2020jaket} introduce descriptive text of entities and their relations into pretraining. Build upon an existing pre-trained encoder\cite{liu2019roberta}, a fully \cite{wang2019kepler} or partially\cite{yu2020jaket} shared encoder is used to encode entity descriptive text with entity-related objectives such as relation type prediction. 

\textbf{Static Entity Representations}
Another way to combine knowledge with a language model is to use static entity representations learned separately from a knowledge base. Ernie (THU)~\cite{zhang2019ernie} and KnowBert~\cite{peters2019knowledge} merge entity representations with language model using entity-to-word attentions. E-Bert~\cite{e_bert} aligns static entity vectors from Wikipedia2Vec with BERT's native wordpiece vector space and uses the aligned entity vectors as if they were wordpiece vectors.

\textbf{Adaptation to Knowledge-Free Model}
It is also possible to incorporate knowledge without joint pretraining or relying on knowledge embeddings. K-BERT~\cite{liu2020k} injects triples from KGs into sentences. A special soft-position and visible matrix in attention are introduced to prevent the injected knowledge from diverting the meaning of the original sentence. K-Adapter~\cite{wang2020k} initializes the model parameters from Roberta~\cite{liu2019roberta} and equips it with adapters to continue training on entity-related objectives.









\section{Approaches}
\label{sec:approaches}

\begin{figure*}[ht]
\resizebox{0.95\textwidth}{!}{%
\includegraphics{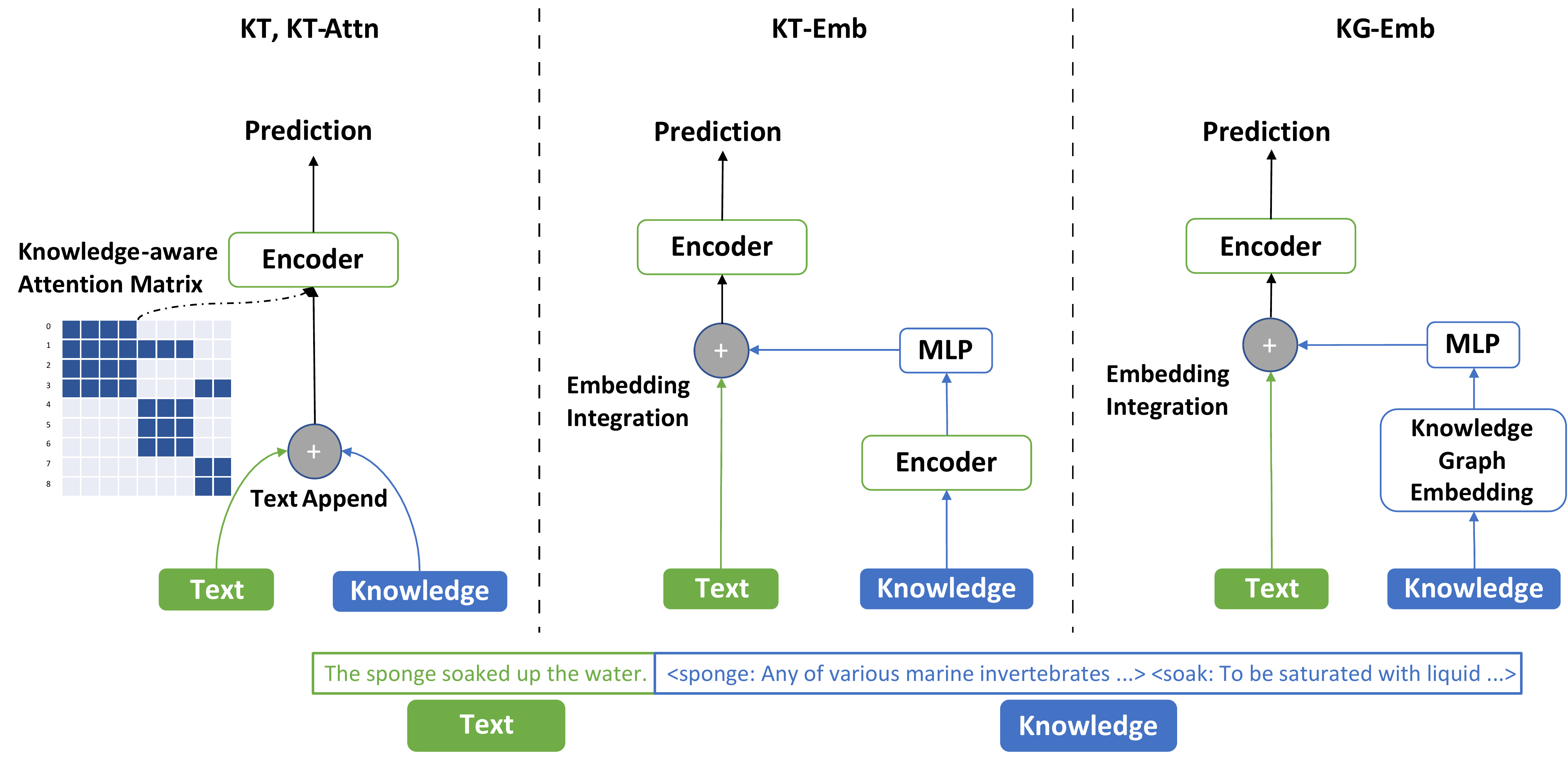}
}
\label{fig:all_methods}
\caption{Illustration of all approaches to incorporate knowledge with language models. Here we assume the encoder module has the flexibility to take either a sequence of tokens or a sequence of token embeddings.
}
\end{figure*}

\subsection{Definition}
Given the input text $x=[x_1, x_2, ..., x_L]$ with $L$ tokens, a language model $f_{LM}$ produces the contextual word representation $f_{LM}(x)=[c_1, c_2, ..., c_L]$. We use $c^{(l)} = [c^{(l)}_1, c^{(l)}_2,..., c^{(l)}_L]$ to represent the intermediate hidden states after $l$ layers. We further assume $c^{(0)}$ represents the token embeddings of $x$. For a specific downstream task, a header function $f_{H}$ further takes the output of $f_{LM}$ as input and generates the prediction as $f_{H}([c_1, c_2, ..., c_L])$.

As we adopt entity information as knowledge in this paper, we assume that the input text $x$ contains $N$ entities $e=[e_1, e_2, ..., e_N]$, where each entity $e_i$ is represented by a contiguous span of tokens in $x$: [$p_i$, $q_i$], where $p_i$ and $q_i$ represents the start and end position of $e_i$. 

\subsection{Combine Knowledge with Language Models}
For a downstream task, the pre-trained language model  $f_{LM}$ and the head function $f_H$ are jointly trained to minimize the loss function on training data $\mathcal{D}$:
\begin{align}
\text{min} \sum_{(x, y)\in \mathcal{D}}L(f_H(f_{LM}(x)), y)\,.
\end{align}
Given the external knowledge $\mathcal{E}$, we explore several general methods to incorporate it into any pre-trained language model $f_{LM}$ such that the knowledge-enhanced language model $f^e_{LM}(x, \mathcal{E})$ can encode the information from both $x$ and $\mathcal{E}$. 

In this work, we consider two formats of knowledge centered on entities:
\begin{itemize}
    \item Free text: An unstructured text $x^{e}$ to describe an entity $e$, e.g. the definition of $e$ from a dictionary;
    \item Embedding: A continuous embedding vector $h^{e}$ to encode an entity $e$, e.g. graph embedding of the node of $e$ from a knowledge graph.
\end{itemize}

To align with the format of knowledge, our integration methods include i) Knowledge as Text, and ii) Knowledge as Embedding, as described in the following sections. 

\begin{table}[t]
\centering
\resizebox{0.5\textwidth}{!}{%
\begin{tabular}{@{}ll@{}}
\toprule
 &
  Example \\ \midrule
Insert after $e_i$ &
  \begin{tabular}[c]{@{}l@{}}The sponge \textit{sponge: Any of various}\\ \textit{marine invertebrates ...} soaked \textit{soak: To}\\ \textit{be saturated with liquid ...} up the water.\end{tabular} \\ \midrule
Append to end &
  \begin{tabular}[c]{@{}l@{}}The sponge soaked up the water. \textit{sponge:} \\ \textit{Any of various marine invertebrates ...}\\ \textit{soak: To be saturated with liquid ...}\end{tabular} \\ \bottomrule
\end{tabular}%
}
\caption{Examples of two approaches of combining external descriptions with text.}
\label{tab:exmp-comb-text}
\end{table}

\subsubsection{Knowledge as Text} The simplest way to incorporate a textual description $x^{e_i}$ with the input $x$ is to concatenate them in the text space. We explore two ways of combination (Table~\ref{tab:exmp-comb-text}): inserting $x^{e_i}$ after $e_i$ in $x$ and appending $x^{e_i}$ to the end of $x$. Empirically we found that the second approach always outperforms the first one in the GLUE benchmark. So we adopt the appending approach as the first knowledge combination method, which we refer to as \textit{Knowledge as Text} (\textbf{KT}). 

As pointed in \citet{liu2020k}, too much knowledge incorporation may divert the sentence from its original meaning by introducing a lot of noise. This is more likely to happen if there are multiple entities in the input text. 
To solve this issue, we adopt the visibility matrix \citep{liu2020k} to limit the impact of descriptions on the original text. In the Transformer architecture, an attention mask matrix is added with the self-attention weights before softmax. Therefore, an $-\infty$ value in attention mask matrix $j,k$ blocks token $j$ from attending to token $k$ and a $0$ value allows token $j$ to attend to token $k$. In our case, we modify the attention mask matrix $M$ such that
\begin{equation}
M_{jk} = 
    \begin{cases}
        0 & x_j, x_k \in x \\
        0 & x_j, x_k \in x^{e_i} \\
        0 & x_j \in x, x_k \in x^{e_i} ~\text{and}~ j = p_i \\
        -\infty & \text{otherwise}
    \end{cases}
\end{equation}
where $x_j$ and $x_k$ are tokens from the concatenation of $x$ and descriptions $[x^{e_1}, x^{e_2}, ..., x^{e_N}]$. In other words, $x_j$ can attend to $x_k$ if: both tokens belong to the input $x$, or both tokens belong to the description of the same entity $e_i$, or $x_j$ is the token at the starting position of entity $e_i$ in $x$ and $x_k$ is from its description text $x^{e_i}$.

\begin{figure}[t]
\centering
\resizebox{0.45\textwidth}{!}{%
\includegraphics{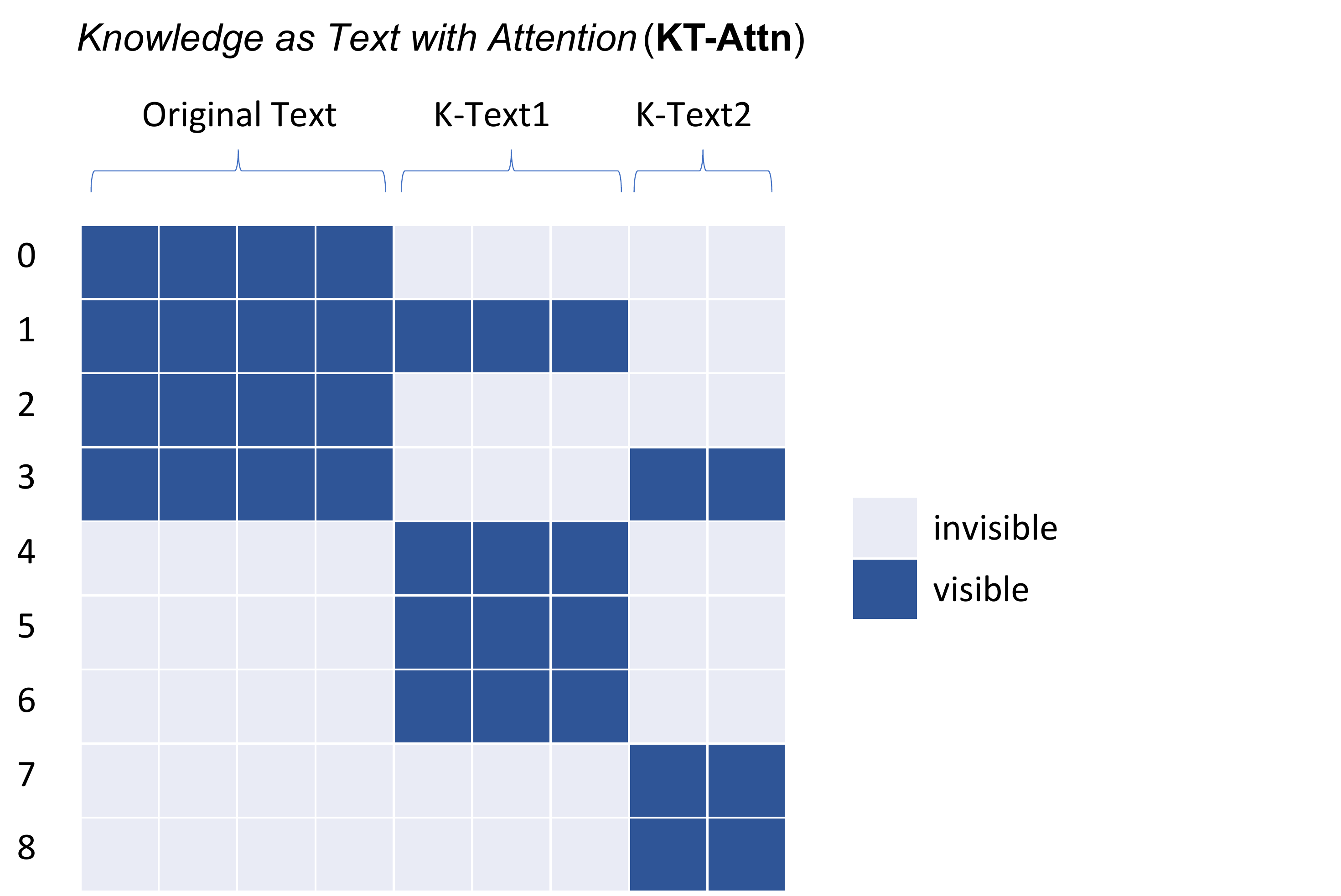}
}
\caption{Illustration of attention matrix in \textbf{KT-Attn}. In this example, K-Text1 describes the token starting at position 1 and K-Text2 describes the token starting at position 3. 
}
\label{fig:kt-attn}
\end{figure}

Figure~\ref{fig:kt-attn} illustrates the attention matrix given a input text and two entity descriptions.
We refer to this approach as \textit{Knowledge as Text with Attention} (\textbf{KT-Attn}).

\textbf{Knowledge as Embedding}. Here, we first represent each entity $e_i$ by an embedding vector $h_i$. When a knowledge graph of entities is available, we can obtain graphical embedding for each entity node. In our experiments, we use the pre-trained TransE \cite{transe} \footnote{TransE embeddings are from \url{http://openke.thunlp.org/}} to get the embedding of each entity in the Wikidata knowledge graph.

We feed $h_i$ into a multi-layer perception layer (MLP) to align with the input embeddings of language models. We then linearly combine the transformed embedding $\text{MLP}(h_i)$ with the input token embedding at position $p_i$. 

As the language model was not exposed to this additional entity embedding during pre-training, we initialize the weight $\alpha$ of $\text{MLP}(h_i)$ to zero and linearly increase its weight during the whole fine-tuning. Define $c^{(0)}$ as the vector representation that is fed into the language model, we have
$$c^{(0)}(x_{p_i}) = \text{embedding\_layer}(x_{p_i}) + \alpha \text{MLP}(h_i) $$
where $\alpha$ is annealed from $0$ to $\lambda\in[0,1]$. We refer to this integration method as \textit{Knowledge as Graph Embedding} (\textbf{KG-Emb}).

When a knowledge graph is not available, we can use entity descriptions $x^{e_i}$ to produce entity embedding $h_i$. Here, we use $f_{LM}$ to encode the entity description $x^{e_i}$ into contextual representation $c^{e_i} = f_{LM}(x^{e_i})$. As shown in Table \ref{tab:exmp-comb-text}, the knowledge text always starts with the token being explained, e.g. \textit{sponge: Any of various marine...}. Therefore we use the contextual representation of the first token in $c^{e_i}$ as entity embedding $h_i$.
We then use $h_i$ in the same way as \textbf{KG-Emb}. Compared with existing work \citep{yu2020jaket,wang2019kepler}, our approach does not require pre-training with external knowledge and can be easily applied to any pre-trained language model in a non-invasive way. We refer to this method as \textit{Knowledge as Textual Embedding} (\textbf{KT-Emb}).

\section{Experiments}
\label{sec:experiments}
In this section, we perform extensive experiments to examine the aforementioned knowledge integration methods in different pre-trained language models (LMs) on a variety of NLU tasks.

\begin{table}[t!]
\centering
\small
\begin{tabular}{lccc}
\toprule
Dataset  & \#Train & \#Val  & Task    \\
\midrule
CoLA  & 8.5K & 1K & regression \\
SST-2  & 67K & 1.8K & classification \\
MNLI & 393K & 20K & classification\\
QQP  & 364K & 391K & classification \\
QNLI & 105K & 4K & classification\\
STS-B & 7K & 1.4K & regression \\
MRPC & 3.7K & 1.7K & classification\\
RTE & 2.5K & 3K & classification\\
POS & 38.2K & 5.5K & sequence labeling \\
NER & 14K & 3.3K & sequence labeling \\
\bottomrule
\end{tabular}
\caption{Statistics of the datasets. \#Train and \#Val are the number of samples for training and validation.
}
\label{tbl:stat}
\end{table}

\begin{table*}[t!]
\centering
\small 
\begin{tabular}{lccccccccc}
\toprule
Model &CoLA & SST-2 & MRPC & STS-B & QQP & MNLI & QNLI & RTE & Avg \\
\midrule
Metrics & Matt. corr. & Acc. &  Acc. & Pear. corr. & Acc. & Acc. & Acc. & Acc. & \\
\midrule
\makecell[l]{RoBERTa-Large\\\cite{liu2019roberta}} & 68.0 & 96.4 & 90.9 & 92.4 & \textbf{92.2} & 90.2 & 94.7 & 86.6 & 88.93 \\
\midrule
RoBERTa-Large (ours)   & 67.02          & 96.22          & 90.93            & 92.71           & 92.15           & 90.59             & \textbf{94.73}    & 90.61          & 89.37 \\
+ KT                   & 68.52	        & 96.33	         & 89.22	        & 92.39	          & 92.01	        & 90.49	            & 94.55	            & 90.97          & 89.31 \\
+ KT-Attn              & \textbf{68.84}	& 96.44	         & \textbf{91.18}	& 92.61	          & 92.09	        & \textbf{90.63}	& 94.67	            & \textbf{91.7}  & \textbf{89.77} \\
+ KT-Emb               & 68.22	        & \textbf{96.56} & 90.69	        & \textbf{92.8}	  & 92.08	        & 90.56	            & \textbf{94.73}	& 90.97          & 89.58 \\
+ KG-Emb               & 68.03	        & 96.44	         & 90.69	        & 92.42 	      & 92.19	        & \textbf{90.63}	& 94.55	            & 90.61          & 89.45 \\
\midrule
RoBERTa-Base (ours)    & 60.07   	    & 94.72	         & \textbf{89.71}	        & 90.95	          & \textbf{91.58}	        & 87.73	            & 92.84	            & 75.09          & 85.34 \\
+ KT                   & \textbf{62.89}	        & 94.72	         & 88.24	        & 89.87	          & 91.57	        & 87.78	            & 92.75	            & 69.68          & 84.69 \\
+ KT-Attn              & 62.35	        & 94.84	         & 89.22	        & \textbf{90.98}           & \textbf{91.58}           & 87.92	            & 92.90	            & \textbf{76.17}          & \textbf{85.75} \\
+ KT-Emb               & 62.43	        & 94.84	         & \textbf{89.71}	        & 90.9	          & 91.49	        & \textbf{88.02}	            & 92.77	            & 73.29          & 85.43 \\
+ KG-Emb               & 61.62	        & \textbf{95.18}          & 88.97 	        & 90.45  	      & 91.5	        & 88.01	            & \textbf{93.06}	            & 73.65          & 85.31 \\
\bottomrule
\end{tabular}
\caption{Results for RoBERTa on classification and regression (CR) tasks. All results are medians over five runs with different seeds on the development set. To validate our results, we follow RoBERTa~\cite{liu2019roberta} to finetune starting from the MNLI model for RoBERTa-large instead of the baseline pretrained model on RTE, STS-B and MRPC tasks. Complete results on other pretrained language models can be found in the Appendix~\ref{sec:all_LMs_results}. 
}
\label{tbl:CR_results}
\end{table*}
\begin{table}[t!]
\centering
\small 
\begin{tabular}{lccc}
\toprule
Model & POS & NER & Avg \\
\midrule
RoBERTa-Large  & 96.95	                 & 96.33                      & 96.64 \\
+ KT           & \textbf{97.06} / 96.93  & 96.21 / \textbf{96.72}     & \textbf{96.89} \\ 
+ KT-Attn      & \textbf{97.06} / 96.94  & 96.18 / 96.32              & 96.69 \\
+ KT-Emb       & 96.98 / 96.97           & 96.67 / 96.62              & 96.83 \\
+ KG-Emb       & 96.95	                 & 96.64                      & 96.80 \\
\midrule
RoBERTa-Base   & 96.88	                 & 95.30                      & 96.09   \\
+ KT           & 97.03 / 96.87           & 95.07 / 95.57              & 96.3  \\
+ KT-Attn      & \textbf{97.05} / 96.87           & 95.21 / 95.3               & 96.18 \\
+ KT-Emb       & 97.03 / 96.91           & 95.70 / 95.69              & \textbf{96.37} \\
+ KG-Emb       & 96.91	                 & \textbf{95.75}                      & 96.33 \\
\bottomrule
\end{tabular}
\caption{Results for RoBERTa on two sequence labeling (SL) tasks. For \textit{KT}, \textit{KT-Attn} and \textit{KT-Emb}, we also experiment with extracting knowledge description for tokens from entity linking which are denoted in right. We report F1 for both tasks. Reported results are medians over five runs on the development set. 
Complete results on other pre-trained language models can be found in the Appendix~\ref{sec:all_LMs_results}.}
\label{tbl:SL_results}
\end{table}

\subsection{Experimental Setup}
Table~\ref{tbl:stat} lists the 10 datasets in our study, including 8 classification and regression (CR) tasks from the GLUE~\cite{glue} benchmark and two sequence labeling (SL) tasks from Penn Treebank~\cite{marcus-etal-1993-building} and CoNLL-2003 shared task data~\cite{conll2003}.
We study on 4 different LMs: $(i)$ RoBERTa~\cite{liu2019roberta}; $(ii)$ BERT~\cite{devlin2019bert}, $(iii)$ ALBERT~\cite{Lan2020ALBERT:} and $(iv)$ ELECTRA~\cite{Clark2020ELECTRA:}. For each language model, we experiment with both base and large models.
Details of datasets and LMs can be found in Appendix~\ref{sec:datasets_and_lms}.

Our implementation is based on HuggingFace's Transformers~\cite{wolf-etal-2020-transformers}. 
We conduct all experiments on 8 Nvidia A100-40GB GPU cards.
We set the fixed training epochs and batch size for each task, and a limited hyperparameter sweep with learning rates $\in$ \{1e-5, 2e-5, 3e-5\}.
For \textit{KT-Emb} and \textit{KG-Emb}, we search warmup weight $\lambda \in$ \{0.1, 0.2, 0.3\}.
For CR tasks, the training epochs are set to 10.
Due to the sufficient training data of MNLI and QQP, we set their epochs to 5.
For SL tasks, we set the training epochs to 3.
The batch size is set to 128 for CR tasks except that we search the batch size in \{16, 32, 128\} for CoLA and STS-B on RoBERTa-base due to their small training data and then fix it for fair comparison~\footnote{CoLA and STS-B use batch size 32 and 16 respectively.}.
For SL tasks, the batch size is set to 16.
We report the median of results on the development set over five fixed random seeds for all tasks.

To extract the knowledge description, we first use Spacy\footnote{https://spacy.io/} to annotate $x$ and select the nouns, verbs, or adjectives to use as the knowledge entities.
For \textit{KG-Emb}, we use REL \cite{van2020rel} to link entities to Wikidata.
We leverage external knowledge source Wiktionary\footnote{https://en.wiktionary.org/wiki/Wiktionary:Main\_Page} to obtain the description for each entity.

\begin{figure}[!ht]

\begin{center}$
\begin{array}{rr}
\includegraphics[width=34mm]{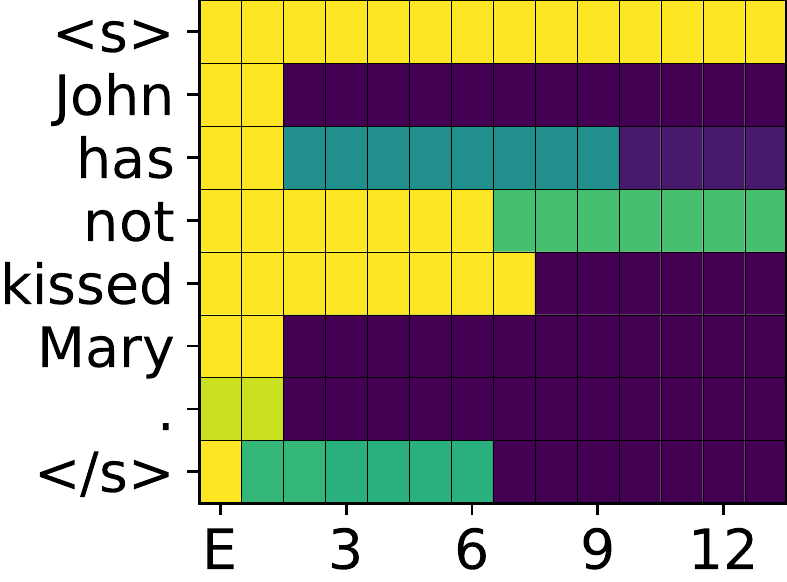}&
\includegraphics[width=34mm]{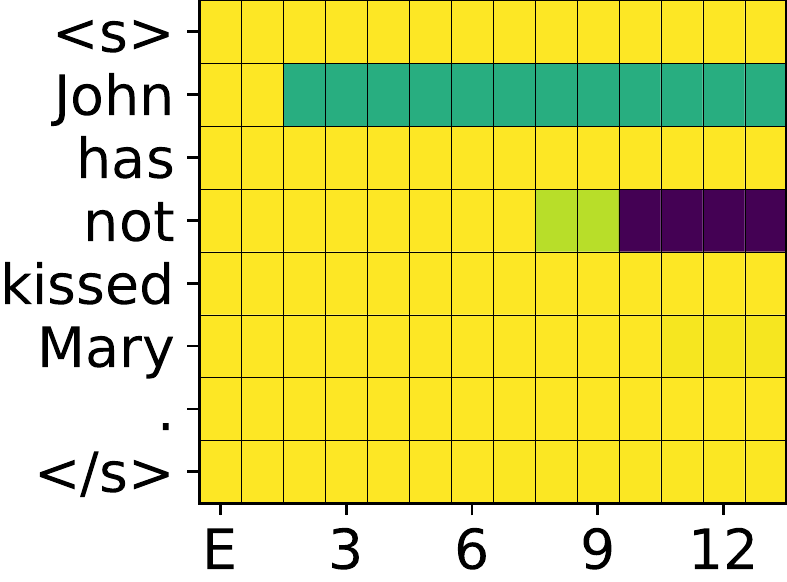}
\end{array}$
\end{center}

\begin{center}$
\begin{array}{rr}
\includegraphics[width=34mm]{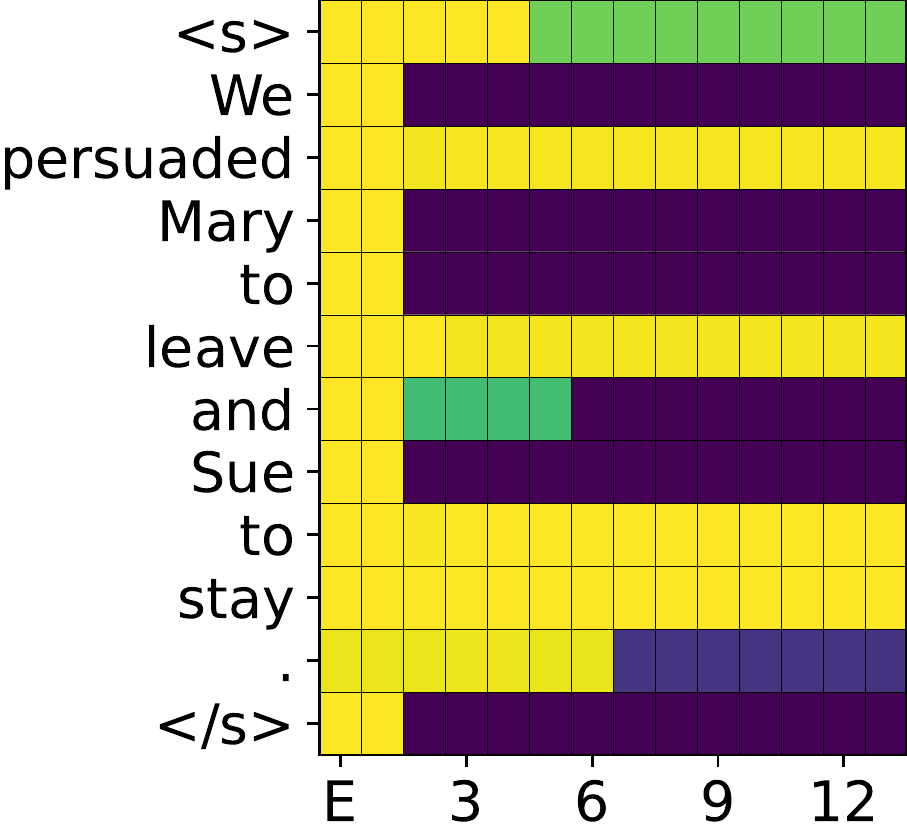} &
\includegraphics[width=34mm]{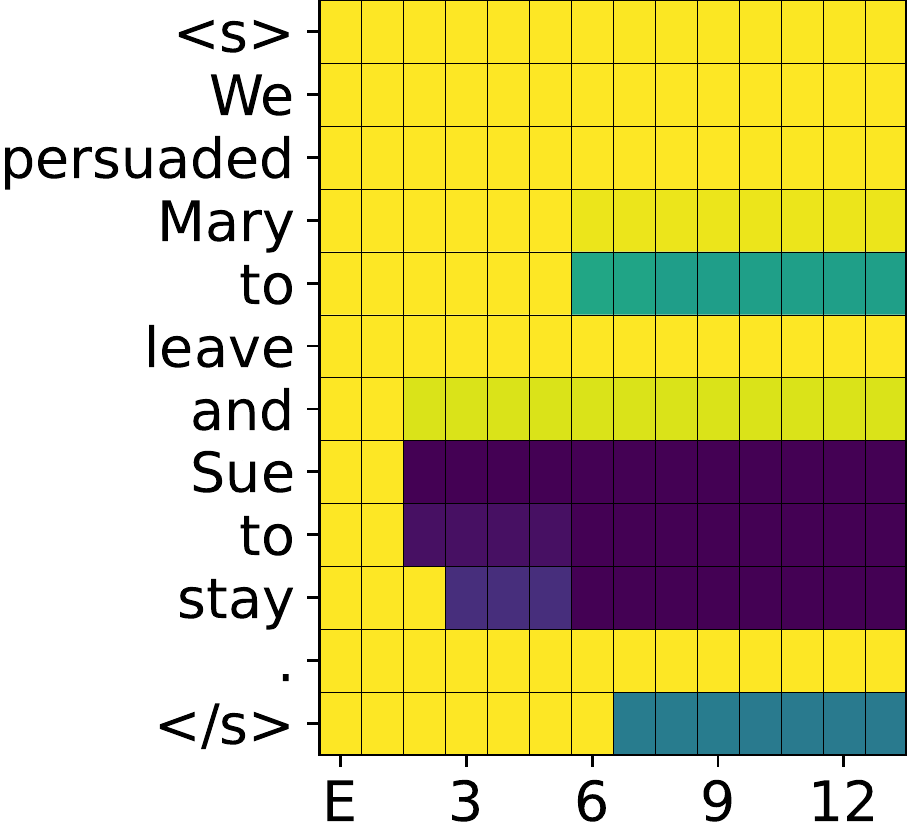}
\end{array}$
\end{center}

\begin{center}$
\begin{array}{rr}
\hspace{0.03\textwidth} \text{(a) RoBERTa} &
\hspace{0.06\textwidth} \text{(b) KT-Attn (Ours)}
\end{array}$
\end{center}

\caption{DiffMask plot for CoLA task with RoBERTa-Base model. CoLA task is to predict the linguistic acceptability of a sentence.
A\colorbox{mypurple}{\textcolor{white}{purple}}cell means that the model's corresponding layer thinks the token on the left is not important for the end task and can be ignored. A\colorbox{myyellow}{yellow}cell means the opposite and green cells mean neutrality. 
The left column shows the result from the vanilla RoBERTa-Base and the right column shows \textbf{KT-Attn} which is one of our knowledge integrated language model. Clearly, KT-Attn has a better understanding of the end task as it correctly identifies words and phrases such as ``not'' and ``and Sue to stay'' which would not change the linguistic acceptability if being ignored.}
\label{fig:diffmask-cola-exmp}
\end{figure}

\subsection{Knowledge Integration Results}
\label{sec:kg_results}
\begin{figure*}[t]

\includegraphics[width=\linewidth]{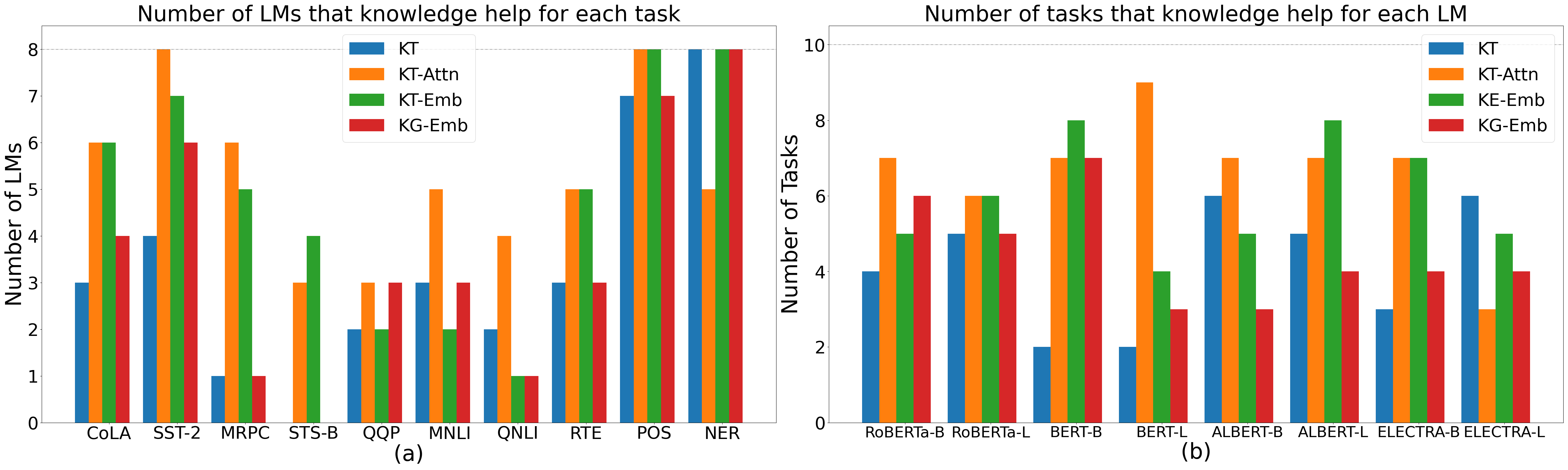}
\includegraphics[width=\linewidth]{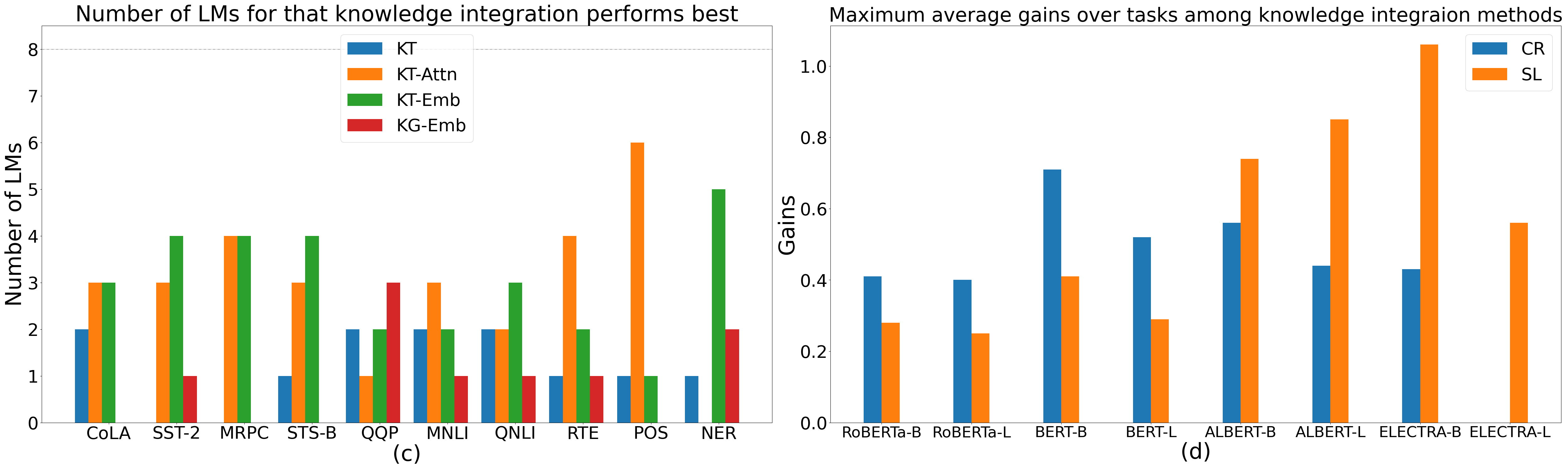}

\caption{Effectiveness of knowledge integration methods on different tasks and language models.
Figure (a) shows for each task the number of language models for which our knowledge integration methods could improve accuracy (maximum is 8 which means it helps all LMs on that task).
Figure (b) shows for each language model the number of tasks that knowledge integration could improve accuracy on (maximum is 10 which means it helps all tasks with that language model).
Figure (c) shows for each task the number of language models that knowledge integration method performs best on (maximum is 10 which means it always performs best with different LMs among 4 integration methods for that task).
Figure (d) shows for each language model the maximum average gains over CR and SL tasks.
`B` and `L` stand for the base and large model respectively. The dashed lines in figure (a), (b) and (c) represent the upper bound. The detailed performance numbers on each task are in the Appendix~\ref{sec:all_LMs_results}.
}
\label{fig:corr_lm_task}
\end{figure*}

In this section, we present different knowledge integration results in 8 pretrained language models.
Table~\ref{tbl:CR_results} and Table~\ref{tbl:SL_results} list detailed numbers on 10 NLU tasks for RoBERTa base and large models. Figure~\ref{fig:corr_lm_task} summarizes our results on all LMs.
From these results, we aim to answer the following questions.

\vspace{5pt}
\noindent\textit{\textbf{Q1: Does knowledge help general NLU tasks?}}
Overall we find that knowledge can help general NLU tasks. Firstly, Table~\ref{tbl:CR_results} shows that \textit{KT-Attn} outperforms both RoBERTa base and large baselines about 0.4 points on average for CR tasks. 
For SL tasks, \textit{KT} and \textit{KT-Emb} outperform baselines about 0.25 and 0.28 points on average.
Secondly, Figure~\ref{fig:corr_lm_task}(a) clearly shows that all tasks can benefit from knowledge across 8 different LMs. For example, \textit{KT-Attn} improves all LMs via the introduction of knowledge for SST-2 and POS tasks.
Thirdly, the average gain of all LMs on 10 NLU tasks with the introduction of knowledge is about 0.46 points. Figure~\ref{fig:corr_lm_task}(d) also shows the average gains with each LMs for CR and SL tasks.

\vspace{5pt}
\noindent\textit{\textbf{Q2: Which tasks benefit the most from knowledge integration?}}
For CR tasks, Table \ref{tbl:CR_results} shows that CoLA, SST2 and RTE get the most improvement.
For SL tasks, Table \ref{tbl:SL_results} shows that both POS and NER get the considerable improvements as regards their strong baselines.
In terms of the number of LMs that knowledge can help with, Figure~\ref{fig:corr_lm_task}(a) shows that SST-2, POS, and NER benefit the most as they get improved on all language models with the introduction of knowledge.

\vspace{5pt}
\noindent\textit{\textbf{Q3: What is the best way to combine KGs with CWR for different NLU tasks?}}
Firstly, Figure~\ref{fig:corr_lm_task}(a) shows that \textit{KT-Attn} and \textit{KT-Emb} can help most LMs for each task.
Secondly, in terms of best knowledge integration methods on each task, Figure~\ref{fig:corr_lm_task}(c) shows that \textit{KT-Attn} and \textit{KT-Emb} accounts more than the other two methods among all LMs.
Thirdly, in terms of which methods to select entities for knowledge extraction, Table~\ref{tbl:CR_results} shows that the POS-based method performs better than entity-linking based for the POS task while it is the opposite for the NER task.

\vspace{5pt}
\noindent\textit{\textbf{Q4: Which large scale pre-trained language models benefit the most from external knowledge?}}
For the number of benefit tasks aspect, Figure~\ref{fig:corr_lm_task}(b) shows that BERT-Large model gets improvement for 9 tasks with \textit{KT-Attn} method.
For the performance gains aspect, Figure~\ref{fig:corr_lm_task}(d) shows that BERT-Base model improves most for CR tasks while ELECTRA-Base model improves most for SL tasks.


\subsection{Analysis}
\label{sec:analysis}
In addition to measuring the performance of knowledge integration methods on NLU tasks, it is also of great value to understand how and why knowledge integration methods help with language models.
In particular, we answer the following two questions.

\vspace{5pt}
\noindent\textit{\textbf{Q1: Is there any indicator to tell knowledge is helpful besides the end-to-end performance?}}
\citet{wang2021revisiting} proposes to enforce local modules to retain as much information about the input as possible while progressively discarding task-irrelevant parts. Inspired by this, we utilize mutual information (MI) to reflect the difference brought by knowledge.

Specifically, we use the mutual information $I(c^{(l)}; x)$ 
to measure the amount of retained information in $l$-th layer about the raw input $x$, and $I(c^{(l)}; y)$ to measure the amount of retained task-relevant information.

We then calculate the difference $\Delta I(c^{(l)}; x)$ and $\Delta I(c^{(l)}; y)$ between knowledge integration methods and baseline for each layer.
If $\Delta I(c^{(l)}; x) > 0$, it means the knowledge integration helps to retain more information about $x$ at layer $l$ than the baseline. If $\Delta I(c^{(l)}; y) > 0$, it means knowledge helps to discard more task-irrelevant information at layer $l$.

To estimate $I(c^{(l)}; x)$, we follow the common practice \cite{vincent2008extracting,rifai2012disentangling} to use the expected error for reconstructing $x$ from $c^{(l)}$ to approximate $I(c^{(l)}; x) \approx \text{max}_{w} [H(x) - R_{w}(x|c^{(l)})]$, where $R_{w}(x|c^{(l)})$ is the reconstruction error and is estimated by masked language modeling to recover the masked tokens, $H(x)$ denotes the marginal entropy of $x$, as a constant.

To estimate $I(c^{(l)}; y)$, we follow \citet{wang2021revisiting} to compute $I(c^{(l)}; y) \approx \text{max}_{\phi} \{ H(y) - \frac{1}{\|D\|} \sum_{(x,y,c^{(l)})\in D} -\text{log} q_{\phi} (y, c^{(l)})  \} $, where $-\text{log} q_{\phi} (y, c^{(l)})$ is the cross-entropy classification loss.

Both the estimations of $I(c^{(l)}; x)$ and $I(c^{(l)}; y)$ require an auxiliary classifier layer connected to each LM Transformer layer's output. We place more details in Appendix~\ref{sec:mi}. 

Figure~\ref{fig:ana_mi} shows the mutual information difference $\Delta I(c^{(l)}; x)$ and $\Delta I(c^{(l)}; y)$ between each KG integration method and the vanilla RoBERTa-base baseline on CoLA dataset.
We observe the following results:
$(i)$ KT and KT-Attn lead to higher $\Delta I(c^{(l)}; x)$ and $\Delta I(c^{(l)}; y)$, indicating that they retain more information about input while discarding task-irrelevant parts. 
$(ii)$ All KG integration methods gradually discard task-irrelevant information and keep more information about output after the first six layers.

\begin{figure*}[t]
\includegraphics[width=\linewidth]{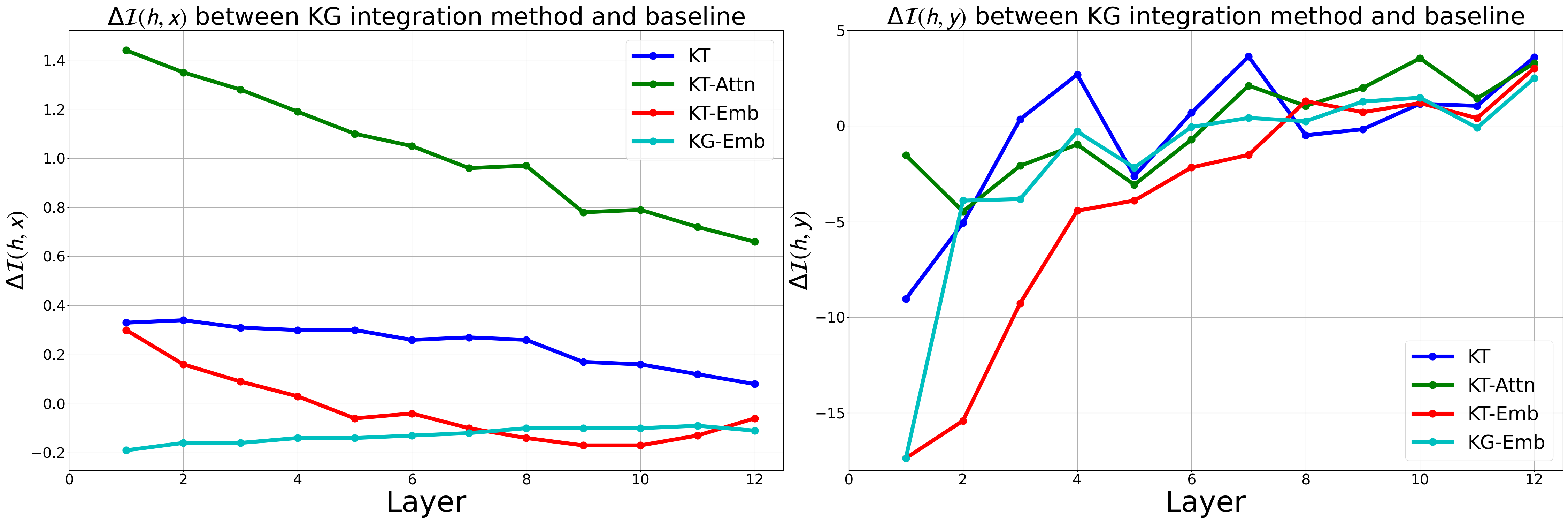}
\caption{
Estimated mutual information difference $\Delta I(h, x)$ and $\Delta I(h, y)$ between each KG integration method and RoBERTa-Base baseline on CoLA.
$\Delta I(c^{(l)}, x) > 0$ means the integration helps retain more information about $x$ at layer $l$ than the baseline. $\Delta I(c^{(l)}, y) > 0$ means that it helps to discard more task-irrelevant information.
}
\label{fig:ana_mi}
\end{figure*}

\begin{figure}[t]

\begin{center}$
\begin{array}{rr}
\includegraphics[width=34mm]{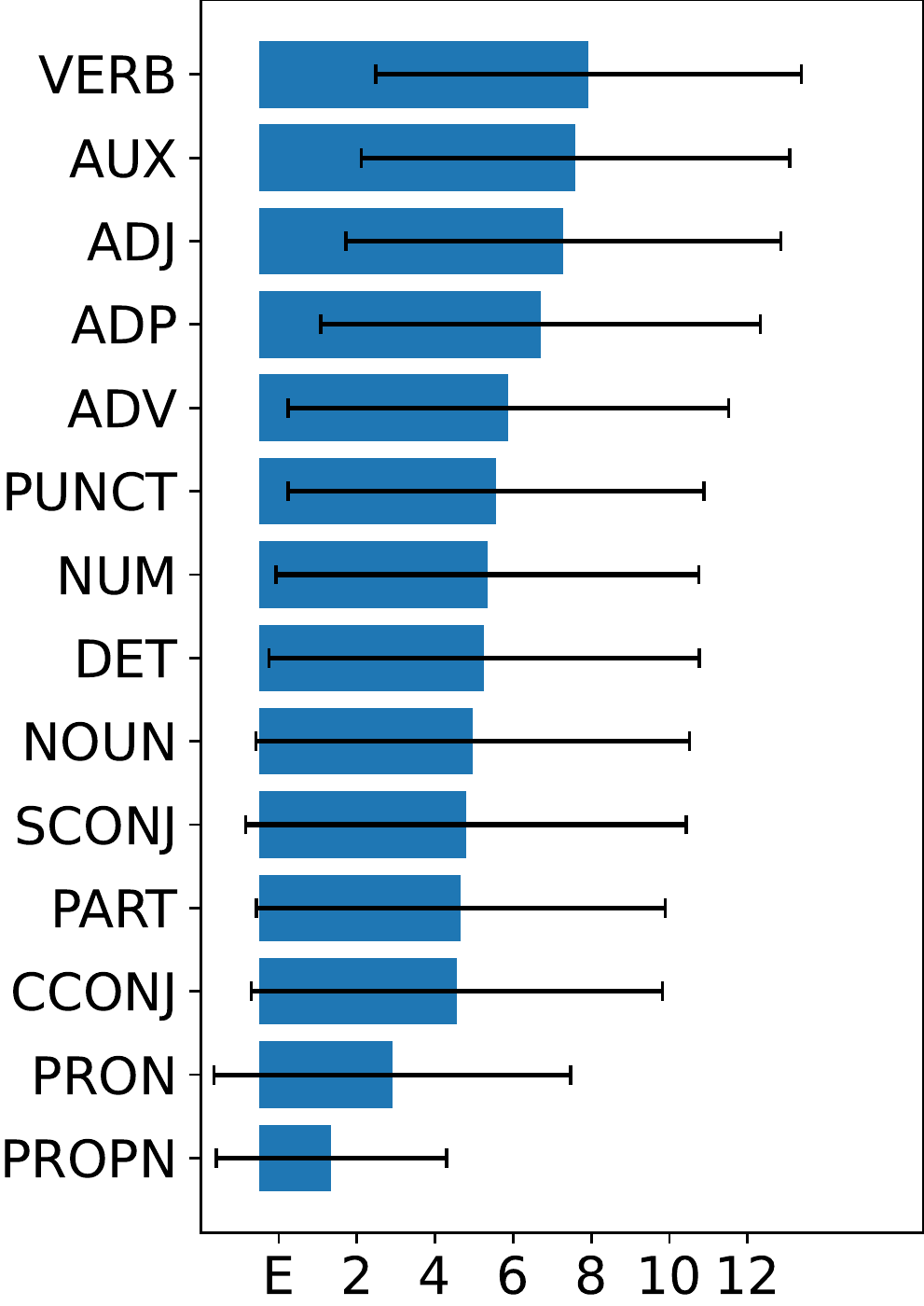} &
\includegraphics[width=34mm]{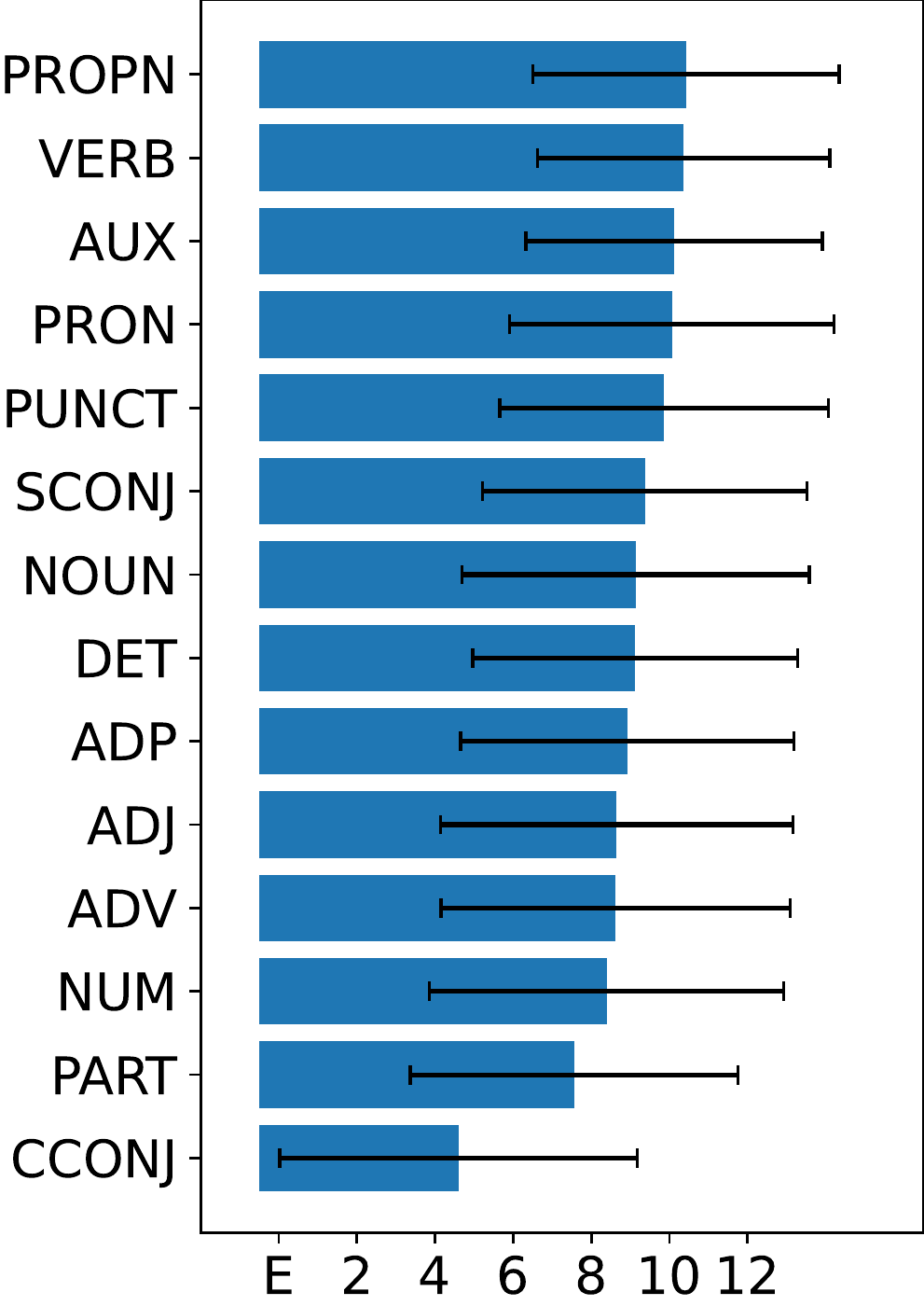}
\end{array}$
\end{center}

\begin{center}$
\begin{array}{rr}
\hspace{0.03\textwidth} \text{(a) RoBERTa} &
\hspace{0.06\textwidth} \text{(b) KT-Attn (Ours)}
\end{array}$
\end{center}

\caption{The average number of Transformer layers in (a) RoBERTa-Base and (b) KT-Attn that deem words of certain part-of-speech as important for the CoLA task of linguistic acceptability. Results are obtained from the DiffMask model \cite{diffmask}.}

\label{fig:diffmask-cola-aggreg-postag}
\end{figure}

\vspace{5pt}
\noindent\textit{\textbf{Q2: How does the introduction of knowledge change the way language models make decisions?}}
We employ DiffMask \cite{diffmask}, an interpretation tool to show how decisions emerge across transformer layers of a language model.
DiffMask learns to mask out subsets of the input while maintaining the output of the network unchanged. The mask value is computed for every token position $j$ at every layer position $l$ by taking the $l$th transformer hidden states $c^{(l)}_j$ as input to an auxiliary classifier. The value of $0$ at token $x_j$ and layer $l$ means the hidden states $c^{(l)}_j$ of token $x_j$ is predictable by the auxiliary classifier that masking $x_j$ in the original input will not affect the model prediction, i.e. the model 'knows' that token $j$ would not influence the final output at layer $l$. A $z_j$ towards $1$ means the opposite. The technical details of DiffMask are described in appendix \ref{subsec:diffmask-detail}.


In Figure \ref{fig:diffmask-cola-exmp} we plot the mask heatmaps of RoBERTa-Base and KT-Attn for two example inputs. As shown, KT-Attn shows better generalization ability since it correctly learns that the negation word in the first example ("not") and the phrase in the second example ("Sue to stay") would not affect the prediction for linguistic acceptability.

In Figure \ref{fig:diffmask-cola-aggreg-postag}, we show the average number of Transformer layers in RoBERTa-Base and KT-Attn that deem words of certain part-of-speech as important for the CoLA task. We can see that although the knowledge is only applied to verbs, nouns, and adjectives, it affects the behavior of the language model on other words as well.
For example, the average number of Transformer layers increases for almost all POS tags of words. And in terms of relative ranking, PRON (pronoun), ADP (adverb), and NUM (numeral) also have significant changes after KT-Attn introduced external knowledge. We include some additional analysis based on DiffMask in appendix \ref{subsec:diffmask-add-analysis}.








\section{Conclusion}
In this paper, we have presented a large-scale empirical study of various knowledge integration methods on 10 general NLU tasks.
We show that knowledge brings more pronounced benefits than previously thought for general NLU tasks since introducing it outperforms across a variety of vanilla pretrained language models and significantly improves the result on certain tasks while having no adverse effects on other tasks. Our analysis with MI and DiffMask further helps understand how and why knowledge integration methods can help with language models.
\bibliographystyle{acl_natbib}
\bibliography{anthology,custom}

\clearpage

\appendix


\begin{center}
    {\Large \bf Appendix}
\end{center}

\section{Datasets and Pretrained Language Models}
\label{sec:datasets_and_lms}
\textbf{CoLA}~\cite{warstadt-etal-2019-neural}: The Corpus of Linguistic Acceptability is a regression dataset annotated about the acceptability whether it is a grammatical English sentence. We use Matthews correlation coefficient~\cite{matthews1975comparison} as the evaluation metric.

\textbf{SST-2}~\cite{socher-etal-2013-recursive}: The dataset of Stanford Sentiment Treebank is a sentiment classification dataset.

\textbf{MRPC}~\cite{dolan-brockett-2005-automatically}: The Microsoft Research Paraphrase Corpus is to predict whether two sentences are semantically equal.

\textbf{STS-B}~\cite{cer-etal-2017-semeval}: The Semantic Textual Similarity Benchmark is the other regression dataset that measures the similarity between the pairs.

\textbf{QQP}\footnote{\url{https://quoradata.quora.com/First-Quora-Dataset-Release-Question-Pairs}}: The Quora Question Pairs is the other dataset to determine whether two sentences are semantically equivalent from the community question-answering website Quora.

\textbf{MNLI}~\cite{williams-etal-2018-broad}: The Multi-Genre Natural Language Inference Corpus is textual entailment tasks and the goal is to classify the relationship between premise and hypothesis sentences into three classes: entailment, contradiction, and neutral.

\textbf{QNLI}~\cite{rajpurkar-etal-2016-squad}: The Stanford Question Answering Dataset is sentence pair classification collection about question-answering. This task is to determine whether the context contains the answer to the question.

\textbf{RTE}~\cite{glue}: The Recognizing Textual Entailment (RTE) datasets is the other textual entailment dataset.

\textbf{POS}: The Part-of-speech tagging is to classify the word in a text to a particular part-of-speech. We use the Penn Treebank~\cite{marcus-etal-1993-building} for this task.

\textbf{NER}: The Name-entity recognition is to seek the name entities among the given sentence. We use the CoNLL-2003 shared task data~\cite{conll2003}.

\begin{table}[t!]
\centering
\begin{tabular}{lc}
\toprule
Models & \#Params \\
\midrule
RoBERTa-Base & 125M \\
RoBERTa-Large & 355M \\
BERT-Base-Cased & 109M \\
BERT-Large-Cased & 335M \\
ALBERT-Base-v2 & 11M \\
ALBERT-Large-v2 & 17M \\
ELECTRA-Base & 110M \\
ELECTRA-Large & 335M \\
\bottomrule
\end{tabular}
\caption{Number of parameters for each pretrained language models in our experiments}
\label{tbl:LMs}
\end{table}
Table~\ref{tbl:LMs} lists all pretrained language models in our experiments.

\section{Knowledge Integration results}
\label{sec:all_LMs_results}
\begin{table*}[t!]
\centering
\small 
\begin{tabular}{lccccccccc}
\toprule
Model & CoLA & SST-2 & MRPC & STS-B & QQP & MNLI & QNLI & RTE & Avg \\
\midrule
Metrics & Matt. corr. & Acc. &  Acc. & Pear. corr. & Acc. & Acc. & Acc. & Acc. & \\
\midrule
BERT-Base-Cased & 60.83 & 92.2 & 84.31 & 89.02 & 90.81 & 83.83 & 91.09 & 66.79 & 82.36 \\
+ KT & 58.38 & 91.86 & 78.68 & 89.01 & 90.88 & 83.83 & 90.76 & 63.18 & 80.82 \\
+ KT-Attn & 59.76 & 92.66 & 85.05 & 89.33 & 90.84 & 83.91 & 90.88 & 66.79 & 82.40 \\
+ KT-Emb & 61.69 & 92.66 & 86.03 & 89.81 & 90.79 & 84.03 & 90.98 & 68.59 & 83.07 \\
+ KG-Emb & 61.25 & 92.43 & 84.56 & 87.88 & 90.83 & 83.9 & 90.92 & 66.43 & 82.28 \\
\midrule
BERT-Large-Cased & 64.84 & 93.92 & 86.27 & 90.29 & 91.5 & 86.51 & 92.51 & 71.48 & 84.67 \\
+ KT & 63.61 & 93.92 & 76.47 & 89.4 & 91.46 & 86.48 & 92.46 & 66.06 & 82.48 \\
+ KT-Attn & 65.35 & 94.04 & 87.25 & 90.56 & 91.43 & 86.63 & 92.57 & 73.65 & 85.19 \\
+ KT-Emb & 64.91 & 93.92 & 87.01 & 90.03 & 91.46 & 86.48 & 92.48 & 71.12 & 84.68 \\
+ KG-Emb & 64.27 & 93.81 & 85.54 & 88.57 & 91.5 & 86.38 & 92.46 & 71.84 & 84.30 \\
\midrule
ALBERT-Base-v2 & 56.31 & 92.83 & 87.75 & 90.88 & 90.63 & 85.13 & 91.76 & 72.2 & 83.44 \\
+ KT & 56.08 & 93.0 & 87.99 & 90.56 & 90.55 & 85.18 & 91.76 & 75.09 & 83.78 \\
+ KT-Attn & 57.42 & 93.0 & 88.48 & 90.81 & 90.54 & 85.08 & 91.91 & 74.73 & 84.00 \\
+ KT-Emb & 55.52 & 93.23 & 88.24 & 90.73 & 90.51 & 85.08 & 91.69 & 74.01 & 83.63 \\
+ KG-Emb & 54.53 & 92.43 & 87.75 & 90.21 & 90.54 & 85.12 & 91.69 & 73.29 & 83.20 \\
\midrule
ALBERT-Large-v2 & 60.16 & 94.38 & 89.22 & 91.39 & 90.88 & 87.18 & 92.59 & 79.42 & 85.65 \\
+ KT & 59.1 & 94.61 & 84.31 & 90.71 & 90.71 & 87.19 & 92.75 & 74.73 & 84.26 \\
+ KT-Attn & 60.55 & 95.07 & 89.71 & 91.21 & 90.9 & 87.1 & 92.62 & 79.42 & 85.82 \\
+ KT-Emb & 61.31 & 94.72 & 89.95 & 91.4 & 90.96 & 87.12 & 92.37 & 80.87 & 86.09 \\
+ KG-Emb & 60.02 & 94.61 & 89.22 & 91.08 & 90.93 & 87.1 & 92.39 & 78.7 & 85.51 \\
\midrule
ELECTRA-Base & 68.61 & 95.3 & 88.48 & 90.99 & 91.93 & 88.9 & 93.1 & 78.7 & 87.00 \\
+ KT & 69.7 & 94.95 & 87.5 & 90.41 & 91.84 & 88.81 & 93.01 & 76.9 & 86.64 \\
+ KT-Attn & 69.45 & 95.76 & 88.97 & 90.89 & 91.84 & 88.98 & 93.03 & 80.51 & 87.43 \\
+ KT-Emb & 70.69 & 95.64 & 88.73 & 91.15 & 91.86 & 88.81 & 93.12 & 76.53 & 87.07 \\
+ KG-Emb & 69.68 & 95.53 & 88.24 & 89.88 & 91.88 & 88.86 & 92.99 & 75.81 & 86.61 \\
\midrule
ELECTRA-Large & 72.13 & 96.67 & 90.93 & 92.57 & 92.45 & 91.32 & 95.15 & 88.45 & 89.96 \\
+ KT & 70.27 & 96.9 & 89.46 & 92.29 & 92.55 & 91.09 & 95.19 & 88.81 & 89.57 \\
+ KT-Attn & 69.25 & 96.79 & 90.44 & 92.23 & 92.58 & 91.28 & 94.98 & 87.73 & 89.41 \\
+ KT-Emb & 68.81 & 97.02 & 90.93 & 91.86 & 92.67 & 91.2 & 95.15 & 88.81 & 89.56 \\
+ KG-Emb & 59.35 & 96.9 & 88.48 & 91.28 & 92.25 & 91.13 & 95.15 & 89.17 & 87.96 \\
\bottomrule
\end{tabular}
\caption{Results on classification and regression (CR) tasks for BERT, ALBERT and ELECTRA.
}
\label{tbl:full_CR_results}
\end{table*}

\begin{table}[t!]
\centering
\small 
\begin{tabular}{lccc}
\toprule
Model & POS & NER & Avg \\
\midrule
Bert-Base-cased & 96.8 & 94.32 & 95.56 \\
KT & 96.77 / 96.79 & 93.82 / 94.46 & 95.62 \\
KT-Attn & 96.93 / 96.79 & 94.21 / 94.33 & 95.63 \\
KT-Emb & 96.9 / 96.82 & 95.02 / 95.03 & 95.97 \\
KG-Emb & 96.83 & 94.94 & 95.88 \\
\midrule
Bert-Large-Cased & 96.85 & 95.39 & 96.12 \\
+ KT & 96.88 / 96.86 & 95.42 / 95.76 & 96.32 \\
+ KT-Attn & 96.99 / 96.86 & 95.53 / 95.44 & 96.26 \\
+ KT-Emb & 96.92 / 96.88 & 95.85 / 95.9 & 96.41 \\
+ KG-Emb & 96.88 & 95.95 & 96.41 \\
\midrule
ALBERT-Base & 96.17 & 93.66 & 94.91 \\
+ KT & 96.76 / 96.23 & 93.91 / 94.4 & 95.58 \\
+ KT-Attn & 96.79 / 96.19 & 94.51 / 93.62 & 95.65 \\
+ KT-Emb & 96.57 / 96.23 & 94.62 / 94.04 & 95.59 \\
+ KG-Emb & 96.22 & 93.75 & 94.98 \\
\midrule
ALBERT-Large & 96.29 & 93.93 & 95.11 \\
+ KT & 96.81 / 96.39 & 94.73 / 94.89 & 95.85 \\
+ KT-Attn & 96.84 / 96.31 & 95.09 / 93.95 & 95.97 \\
+ KT-Emb & 96.73 / 96.34 & 95.17 / 94.45 & 95.95 \\
+ KG-Emb & 96.34 & 94.43 & 95.39 \\
\midrule
ELECTRA-Base & 96.35 & 94.09 & 95.22 \\
+ KT & 96.77 / 96.37 & 94.91 / 94.58 & 95.84 \\
+ KT-Attn & 96.8 / 96.34 & 94.79 / 94.25 & 95.80 \\
+ KT-Emb & 96.86 / 96.49 & 95.71 / 94.89 & 96.28 \\
+ KG-Emb & 96.47 & 94.92 & 95.69 \\
\midrule
ELECTRA-Large & 96.55 & 95.32 & 95.94 \\
+ KT & 96.9 / 96.56 & 95.67 / 95.8 & 96.35 \\
+ KT-Attn & 96.7 / 96.58 & 95.15 / 95.21 & 95.95 \\
+ KT-Emb & 96.86 / 96.64 & 96.14 / 95.72 & 96.50 \\
+ KG-Emb & 96.57 & 95.51 & 96.04 \\
\bottomrule
\end{tabular}
\caption{Results on sequence labeling (SL) tasks for BERT, ALBERT and ELECTRA.}
\label{tbl:full_SL_results}
\end{table}

Table~\ref{tbl:full_CR_results} and Table~\ref{tbl:full_SL_results} list detailed numbers for CR and SL tasks on BERT, ALBERT and ELECTRA.

\section{Mutual Information Implementation Details}
\label{sec:mi}
In our implementation, we stack one transformer layer followed by two fully-connected layers $\phi$ on top of the intermediate hidden states $c^{(l)}$ and optimize the newly added transformer to predict the label $y$. Follow~\cite{wang2021revisiting}, we simply use test accuracy as the estimate of of $I(c^{(l)}; y)$.

\section{DiffMask}
\label{sec:diffmask-appd}

\subsection{Implementation Details}
\label{subsec:diffmask-detail}
DiffMask attaches an MLP classifier to each LM layer's output, including the token embedding layer as layer $0$. The $l$-th classifier $g_{\phi}^{(l)}$ takes the hidden states up to the $l$-th layer to predict a binary mask vector: $v^{(l)} = g_{\phi}^{(l)} (c^{(0)}, ..., c^{(l)})\in \{0, 1\}^n$, where $n$ is the number of input tokens. 

Then, the token mask for each input token $x_j$ is defined as the product of all binary masks up to the $l$-th layer: $z^{(l)}_{j} = \prod_{k=0}^{l} v_j^{(k)} $. The embedding of the masked token is replaced by a learned baseline vector $b$, i.e. $\hat{c}_j^{(0)} = z^{(l)}_j \cdot c_i^{(0)} + (1 - z_i) \cdot b$. The masked embeddings $\hat{c}$ is input the to the finetuned model to get $f_H(f_{LM}(\hat{c}))$. Here we assume $f_{LM}$ could either take tokens $x$ or token embeddings $c$ as input.
The objective of DiffMask is to estimate the parameters of the masking networks and the baseline $b$ to mask-out as many input tokens as possible while keeping $f_H(f_{LM}(x)) \approx f_H(f_{LM}(\hat{c}))$, i.e. keeping the output of masked tokens close to the original output without masks.

According to \citet{diffmask}, the learned masks $z^{(l)}_j$ reveal what the network ``knows'' at layer $l$ about the NLU task. We can therefore plot a heatmap over $\{z^{(l)}_j\}_{l=0,j=1}^{L,n}$. If $z^{(l)}_j=0$, it means that masking the $j$-th input token will not affect the model prediction, i.e. the model 'knows' that token $j$ would not influence the final output at layer $l$ and higher.

\subsection{Additional Analysis}
\label{subsec:diffmask-add-analysis}

\begin{figure}[ht]
\begin{center}$
\begin{array}{rrr}
\includegraphics[width=22mm]{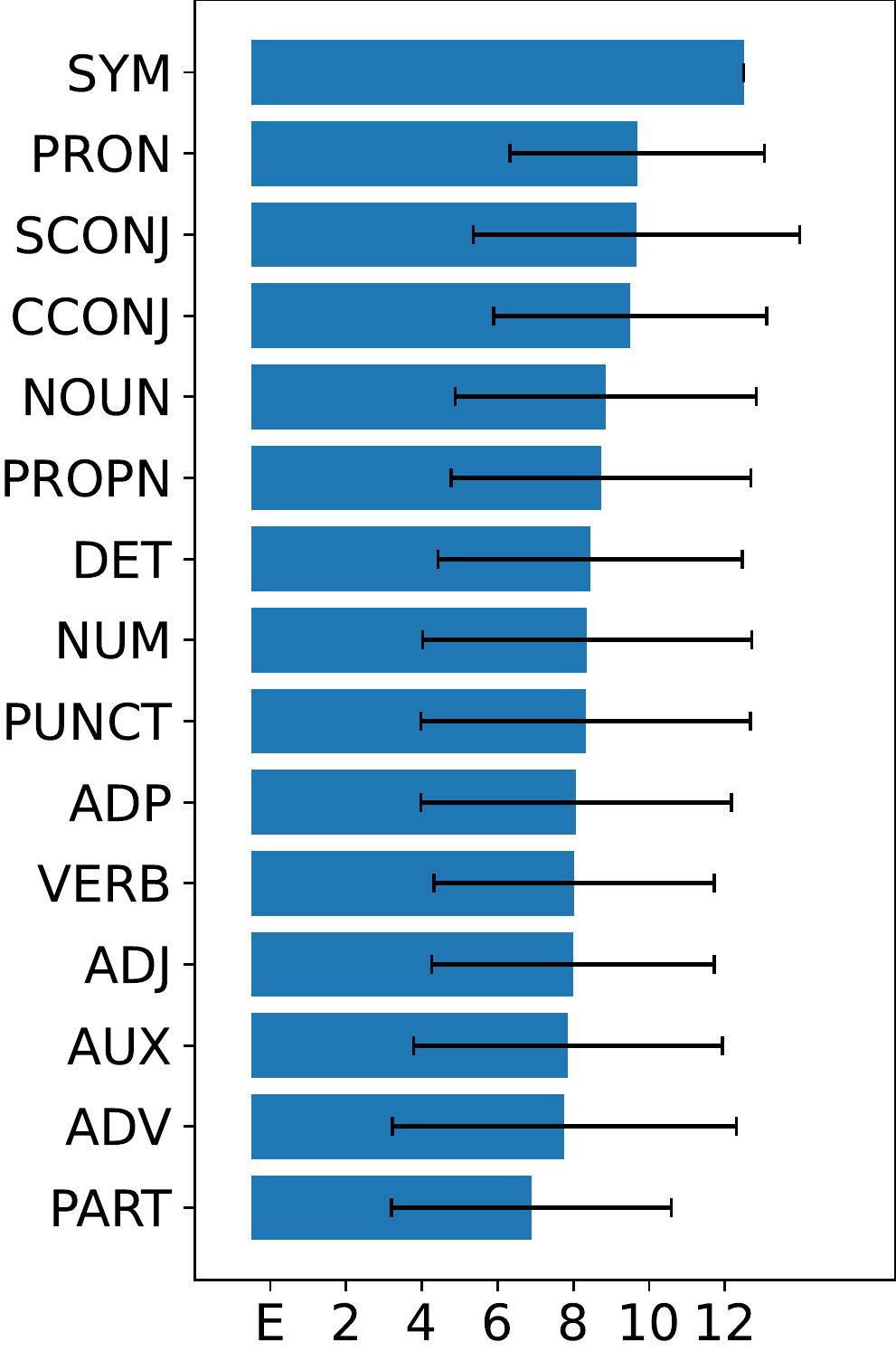} &
\includegraphics[width=22mm]{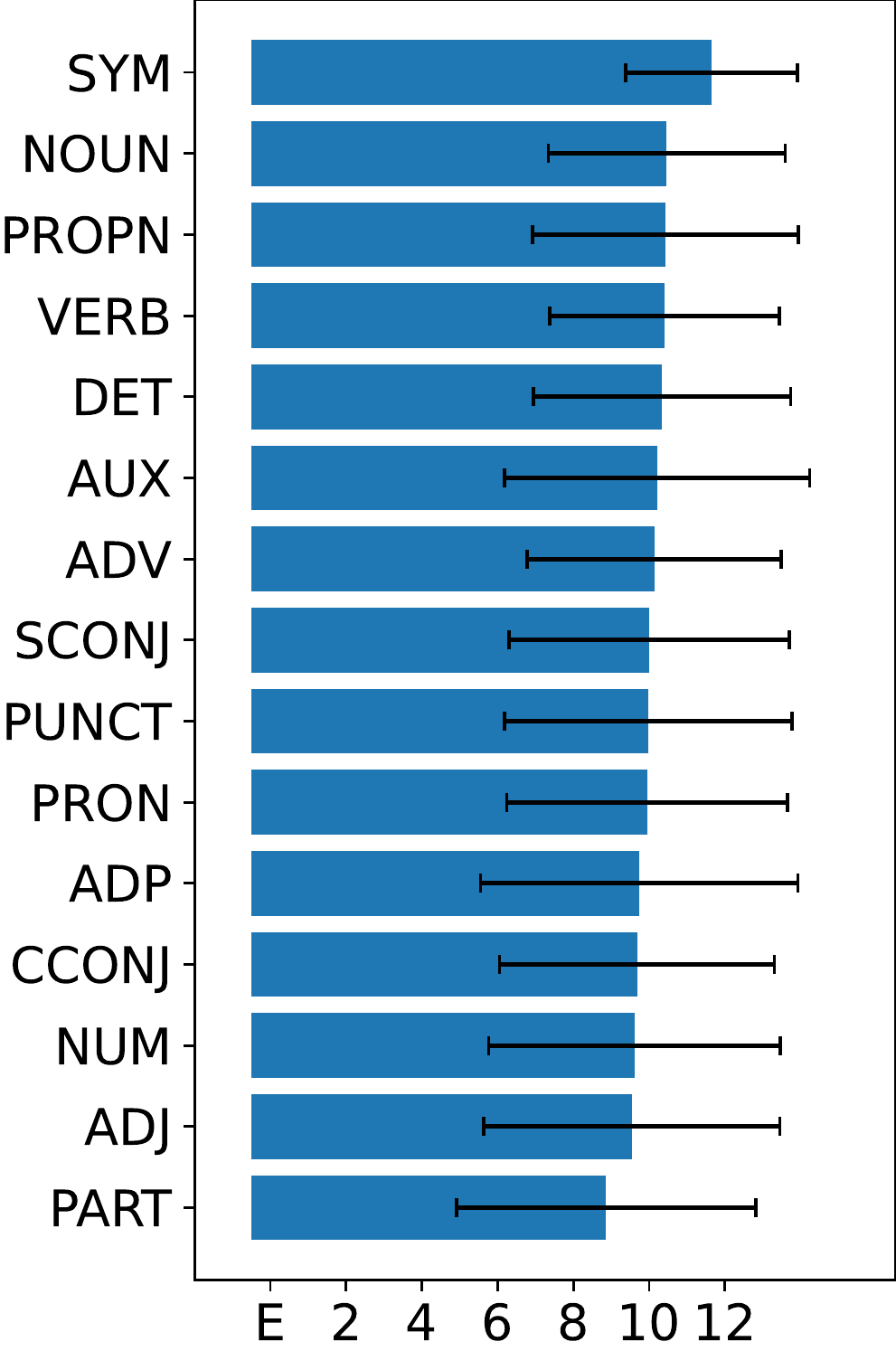} &
\includegraphics[width=22mm]{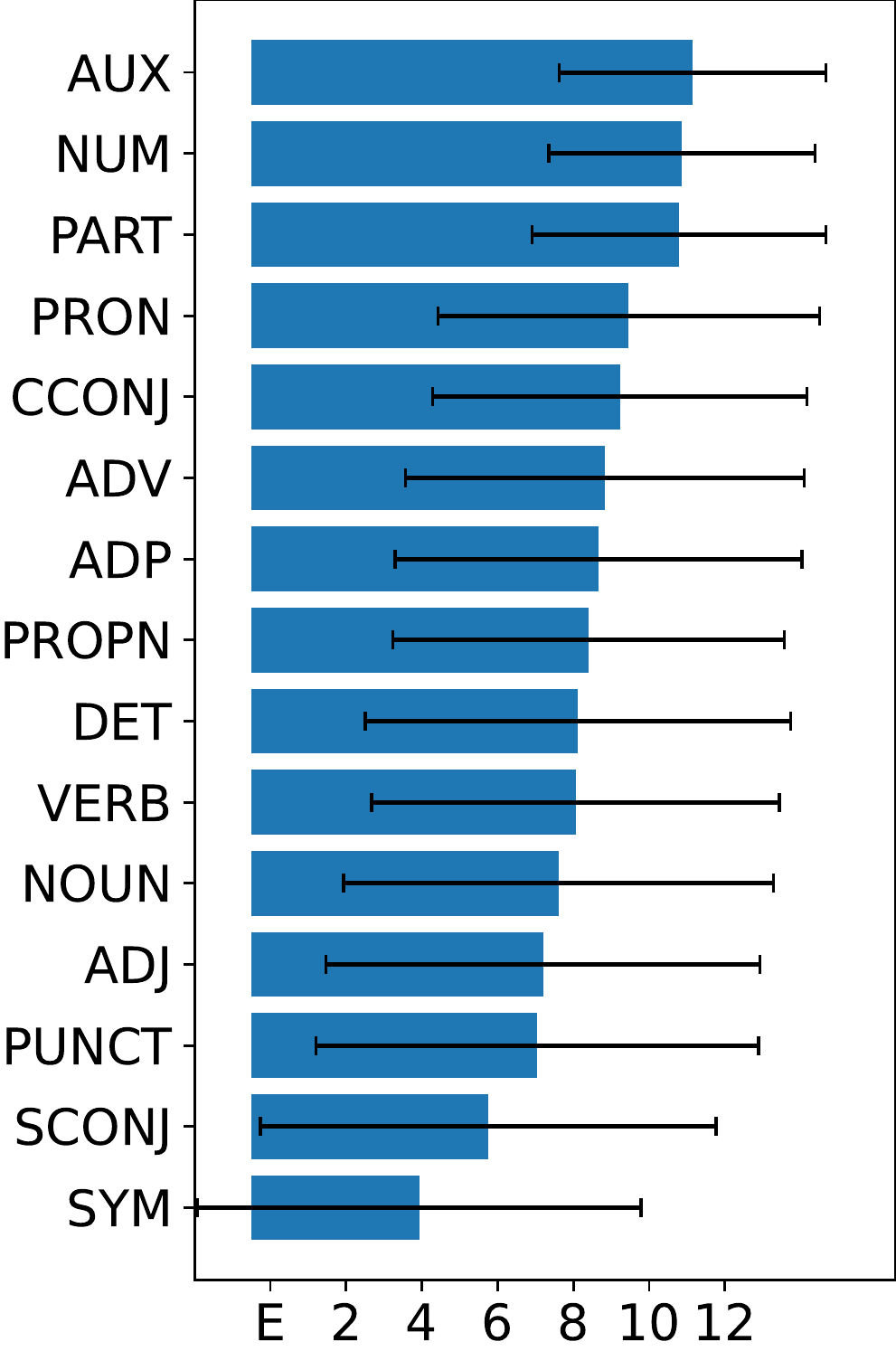}
\end{array}$
\end{center}

\begin{center}$
\begin{array}{rrr}
\hspace{0.02\textwidth} \text{(a) RoBERTa} &
\hspace{0.02\textwidth} \text{(b) KT-Attn} &
\hspace{0.02\textwidth} \text{(c) KT-Emb}
\end{array}$
\end{center}

\caption{The average number of Transformer layers in (a) RoBERTa-Base and (b) KT-Attn (c) KT-Emb that deem words of certain part-of-speech as important for the RTE task. Results are  obtained from the DiffMask model \cite{diffmask}.}

\label{fig:diffmask-rte-aggreg-postag}
\end{figure}

\begin{figure*}[ht]
\begin{center}$
\begin{array}{rr}
\includegraphics[width=40mm]{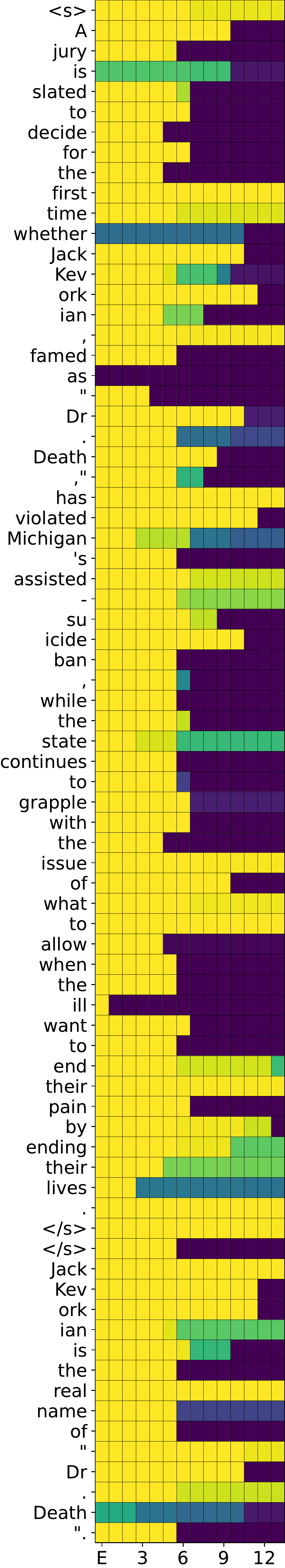} &
\includegraphics[width=40mm]{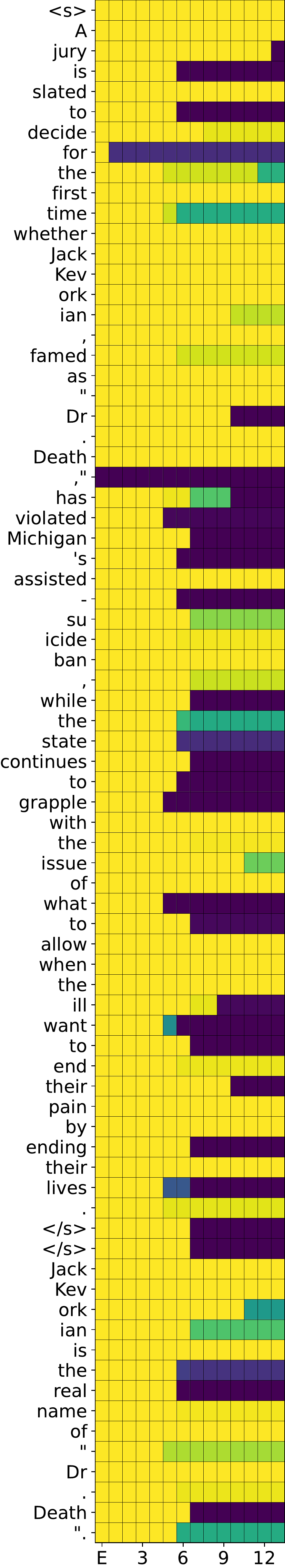}
\includegraphics[width=40mm]{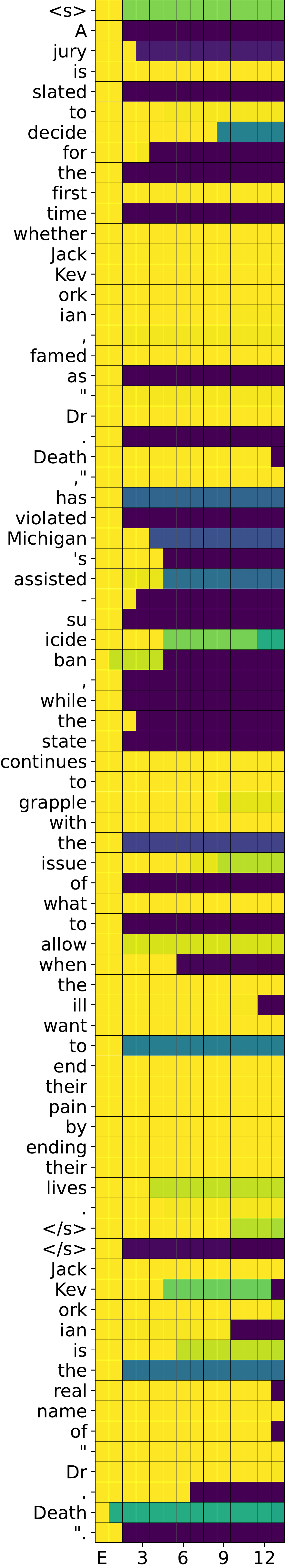}
\end{array}$
\end{center}

\begin{center}$
\begin{array}{rr}
\hspace{0.09\textwidth} \text{(a) RoBERTa} &
\hspace{0.12\textwidth} \text{(b) KT-Attn (Ours)}
\hspace{0.09\textwidth} \text{(c) KT-Emb (Ours)}
\end{array}$
\end{center}

\caption{DiffMask plot for RTE task with RoBERTa-Base model. RTE task is to predict whether two sentences entail each other.
}
\label{fig:diffmask-rte-exmp}
\end{figure*}

In figure \ref{fig:diffmask-rte-exmp}, we plot the DiffMask heatmaps of an example input sentence in the RTE text entailment task. Given two sentences concatenated into a single sequence, the language model RoBERTa-Base is finetuned to predict whether the two sentences entail each other or not. From this example, we can see that the first sentence is verbose while the second one is concise. Therefore, as for entailment judgment, a model with good generalization power should focus on the tokens containing the key information: "Jack Kevorkian", "famed as", "real name" and "Dr. Death". KT-Attn and KT-Emb rely more on those key information than vanilla RoBERTa. In figure \ref{fig:diffmask-rte-aggreg-postag}, we could also see the difference made by introducing knowledge into the finetuning of the language model is not limited to the tokens where knowledge is explicitly incorporated.

\begin{figure}[ht]
\begin{center}$
\begin{array}{rrr}
\includegraphics[width=22mm]{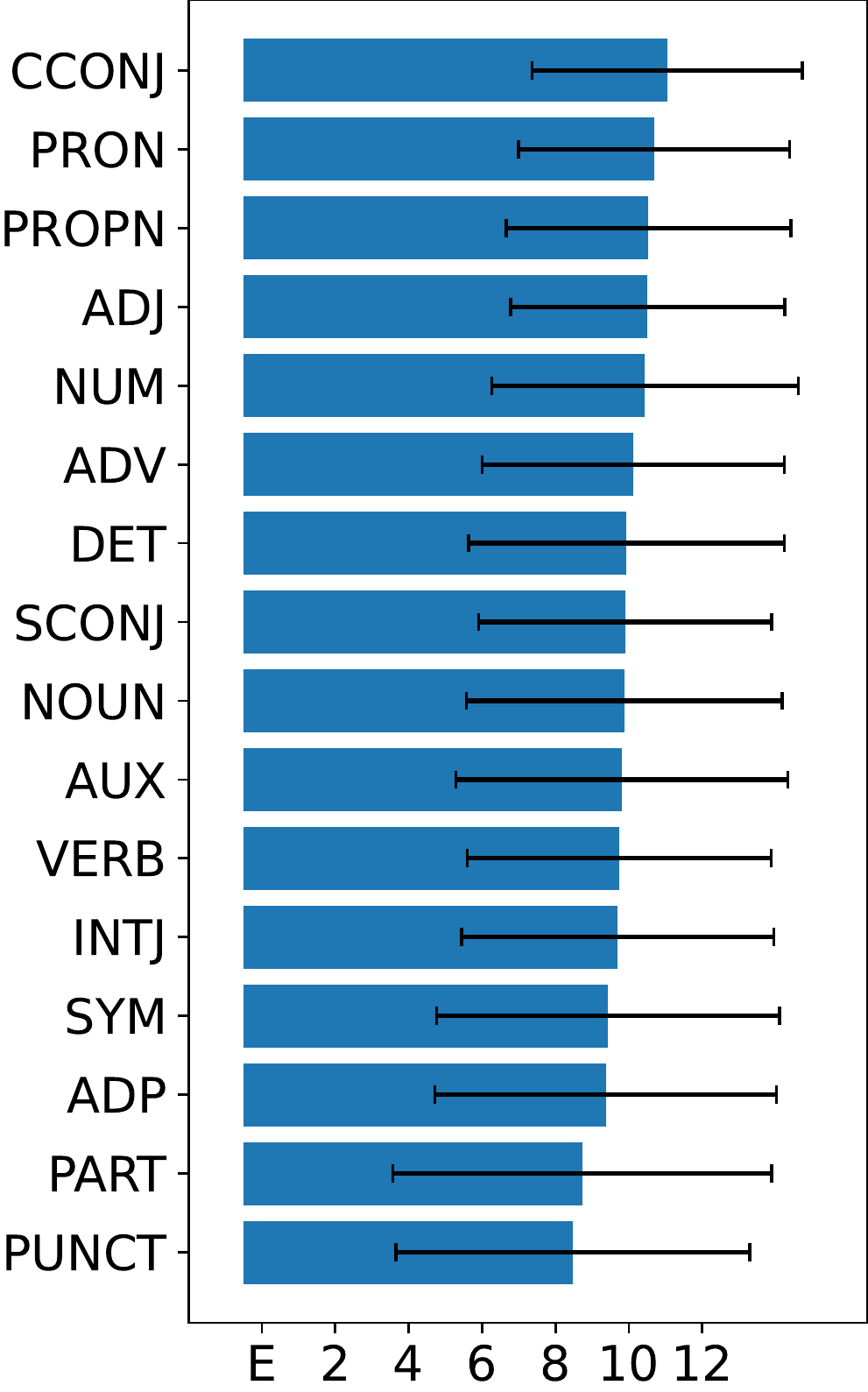} &
\includegraphics[width=22mm]{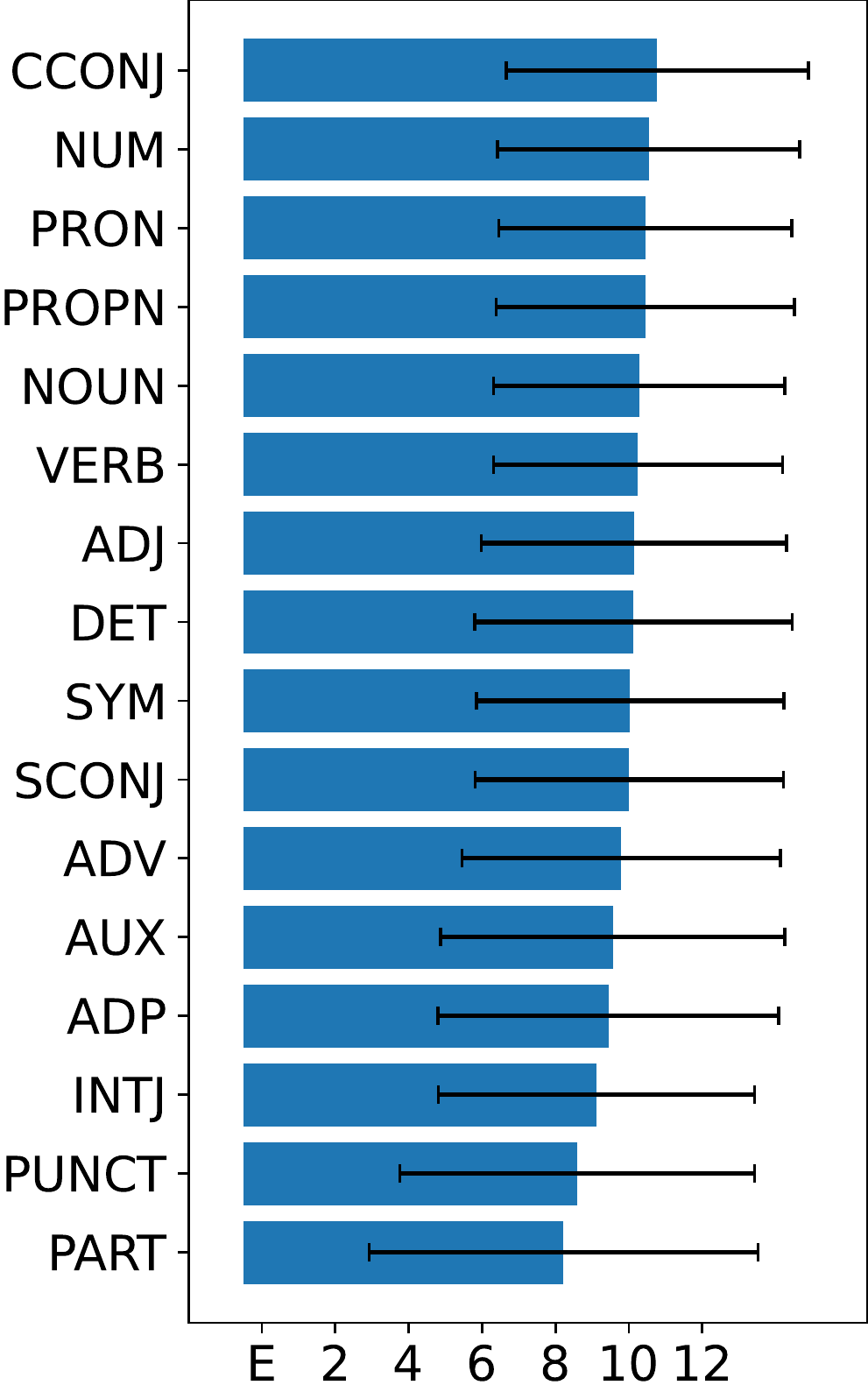} &
\includegraphics[width=22mm]{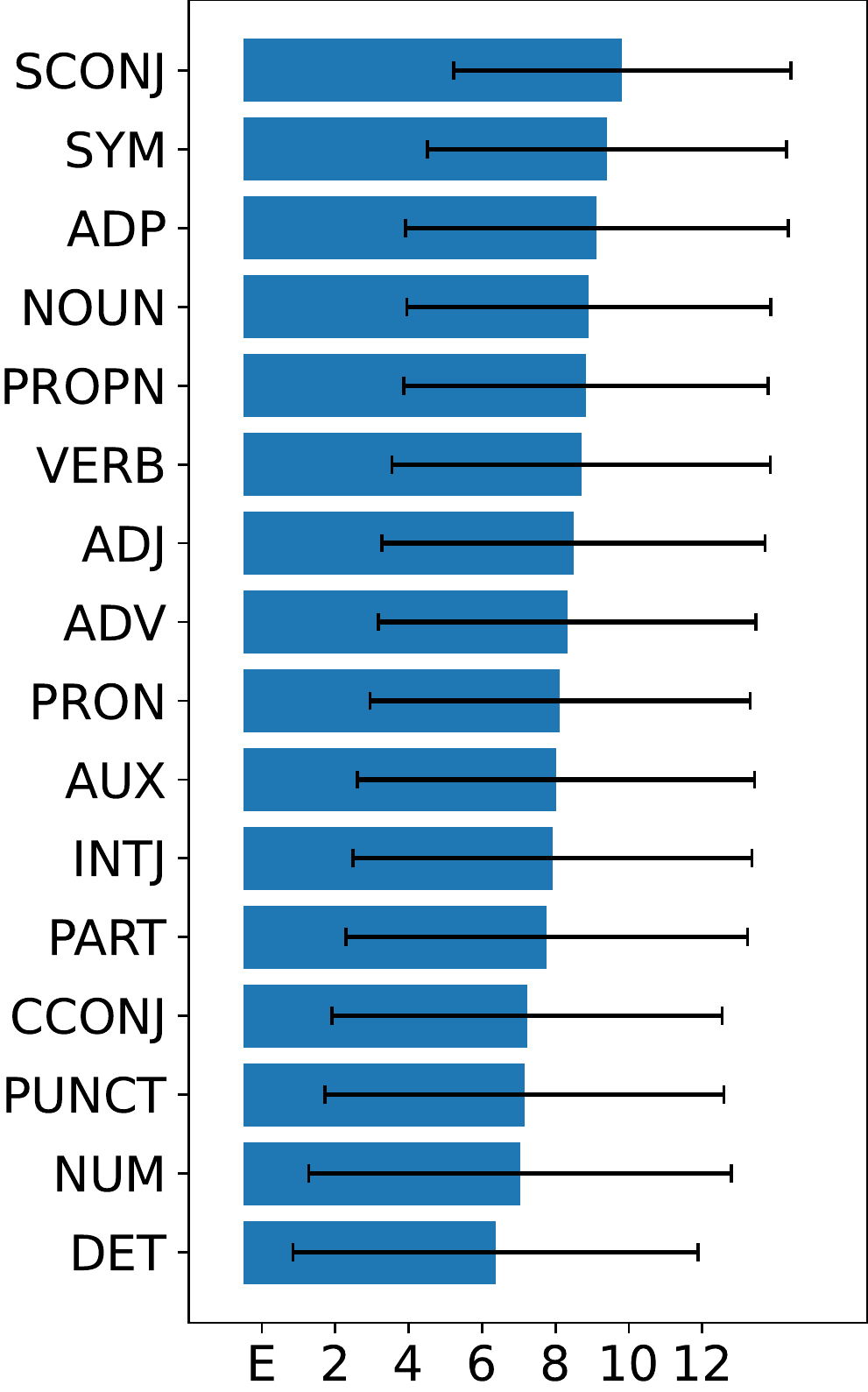}
\end{array}$
\end{center}

\begin{center}$
\begin{array}{rrr}
\hspace{0.02\textwidth} \text{(a) RoBERTa} &
\hspace{0.02\textwidth} \text{(b) KT-Attn} &
\hspace{0.02\textwidth} \text{(c) KT-Emb}
\end{array}$
\end{center}

\caption{The average number of Transformer layers in (a) RoBERTa-Base and (b) KT-Attn  (c) KT-Emb that deem words of certain part-of-speech as important for the STSB task. Results are  obtained from the DiffMask model \cite{diffmask}.}

\label{fig:diffmask-stsb-aggreg-postag}
\end{figure}

\begin{figure*}[t]
\begin{center}$
\begin{array}{rr}
\includegraphics[width=40mm]{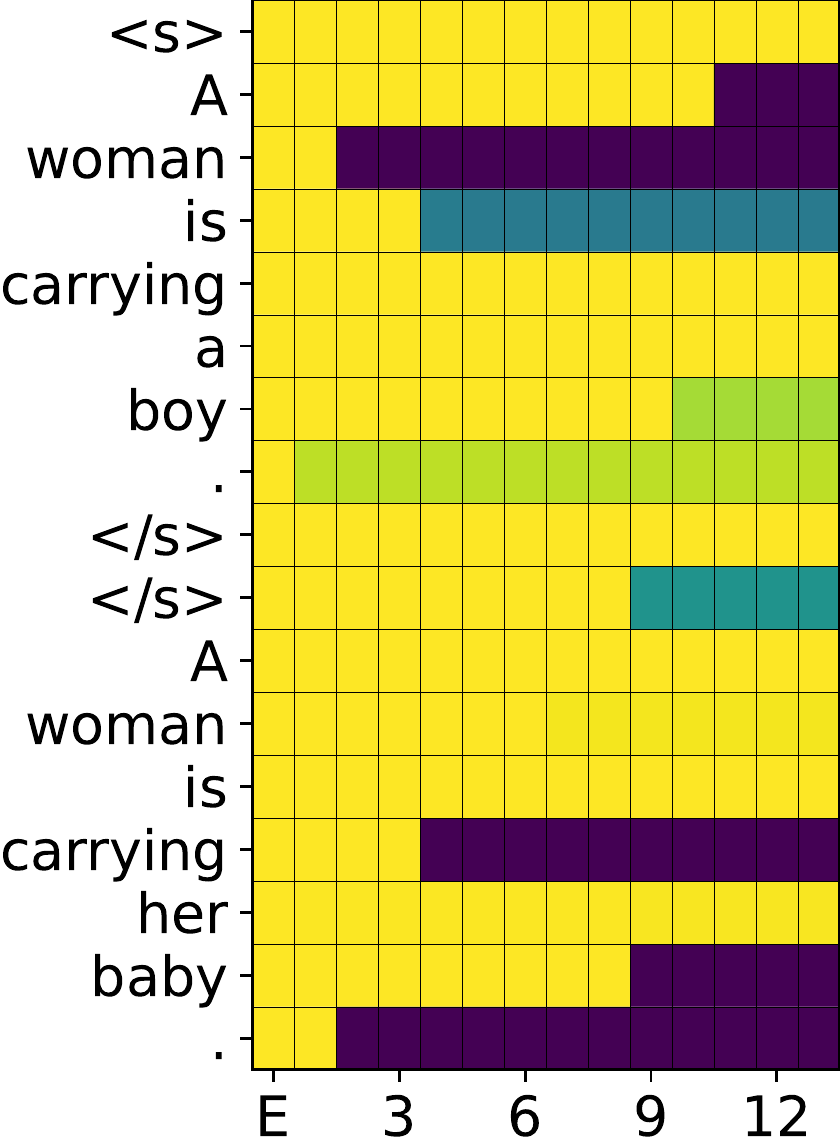} &
\includegraphics[width=40mm]{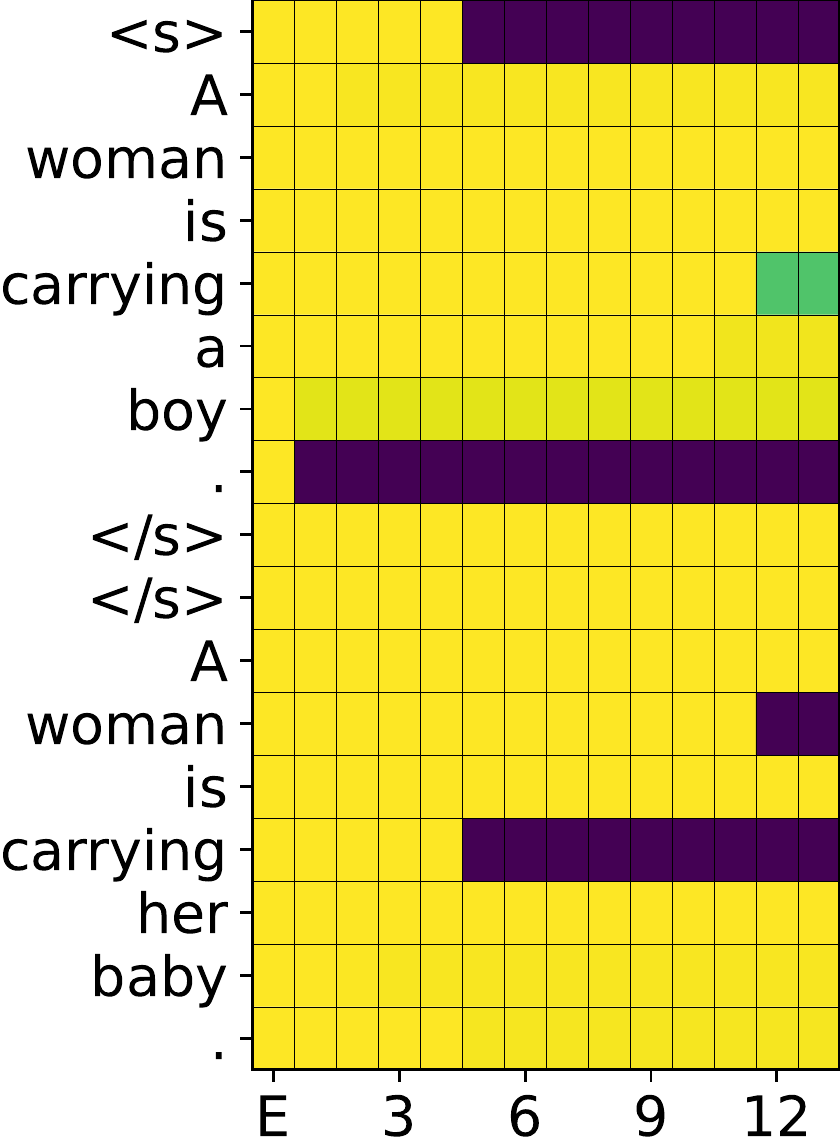}
\includegraphics[width=40mm]{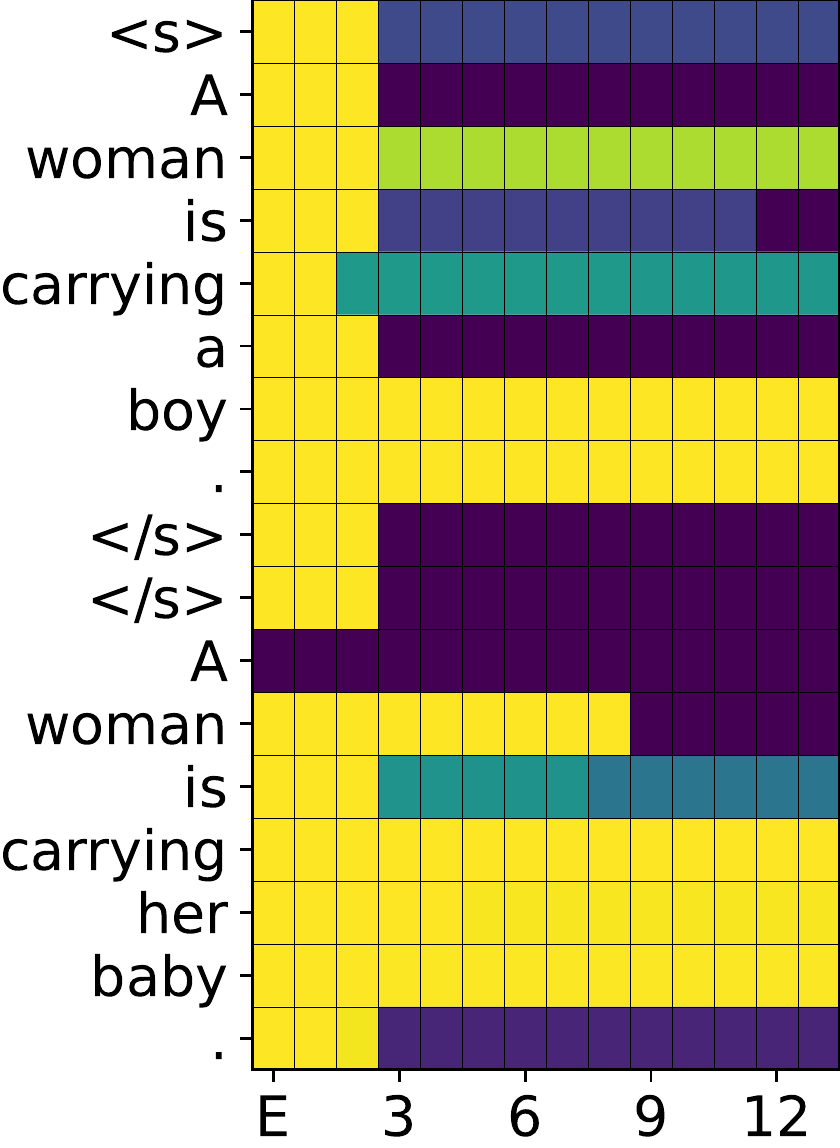}
\end{array}$
\end{center}

\begin{center}$
\begin{array}{rr}
\hspace{0.09\textwidth} \text{(a) RoBERTa} &
\hspace{0.12\textwidth} \text{(b) KT-Attn (Ours)}
\hspace{0.09\textwidth} \text{(c) KT-Emb (Ours)}
\end{array}$
\end{center}

\caption{DiffMask plot for STSB task with RoBERTa-Base model. STSB task is to predict the semantic textual similarity of two sentences.
}
\label{fig:diffmask-stsb-exmp}
\end{figure*}

For STSB, where incorporating knowledge did not show significant improvement of end-to-end performance, we plot one of the examples in figure \ref{fig:diffmask-stsb-exmp} and the average number of Transformer layers that deem words of certain part-of-speech as important for the STSB task in figure \ref{fig:diffmask-stsb-aggreg-postag}. In figure \ref{fig:diffmask-stsb-exmp}, KT-Attn and KT-Emb still show better generalization ability by identifying the keywords "a boy" and "her baby" better than the vanilla RoBERTa model. But the difference is slim since the vanilla RoBERTa model also captures "a" and "her" as the evidence for the final prediction. In figure \ref{fig:diffmask-stsb-exmp}, we can observe a smaller difference between vanilla RoBERTa and its two knowledge-enhanced versions, which indicates that the language models adapt to external knowledge less aggressively for some certain tasks than the others.

\end{document}